\documentclass[10pt]{article} % For LaTeX2e
\usepackage[preprint]{rlc}

\usepackage{amssymb}            % Defines common symbols like \mathbb R
\usepackage{mathtools}          % Extends amsmath, providing common math tools
\usepackage{mathrsfs}           % Enables \mathscr, which can work in cases that \mathcal does not
\mathtoolsset{showonlyrefs}     % Only number equations that are referenced (optional)
\usepackage{graphicx}           % For including images
\usepackage{subcaption}         % Allows for the use of subfigures and subcaptions
\usepackage[space]{grffile}     % For spaces in image names
\usepackage{url}                % For displaying urls

\usepackage{amsmath}
\usepackage{algorithm}
\usepackage{algpseudocode}
\usepackage{comment}
\usepackage{multirow}

\usepackage{cleveref}
\usepackage{bm}
\usepackage{booktabs}
\usepackage{wrapfig}

\newcommand{\MDP}[0]{M}                       % MDP (Markov Decision Process)
\newcommand{\statespace}[0]{\mathcal{S}}                % State space
\newcommand{\actionspace}[0]{\mathcal{A}}               % Action space
\newcommand{\transdomain}[0]{\mathcal{T}}               % Transition domain (probability distribution of algorithm state transitions)
\newcommand{\rewards}[0]{\mathcal{R}}                   % Reward function
\title{Inferring Behavior-Specific Context Improves Zero-Shot Generalization in Reinforcement Learning}

\author{Tidiane Camaret Ndir\hspace{27.1mm} André Biedenkapp\hspace{27.1mm} Noor Awad\\
\{ndirt, biedenka, awad\}@cs.uni-freiburg.de\hfill \hfill University of Freiburg}

\begin{document}

\maketitle

\begin{abstract}

In this work, we address the challenge of zero-shot generalization (ZSG) in Reinforcement Learning (RL), where agents must adapt to entirely novel environments without additional training. We argue that understanding and utilizing contextual cues, such as the gravity level of the environment, is critical for robust generalization, and we propose to integrate the learning of context representations directly with policy learning. Our algorithm demonstrates improved generalization on various simulated domains, outperforming prior context-learning techniques in zero-shot settings. By jointly learning policy and context, our method acquires behavior-specific context representations, enabling adaptation to unseen environments and marks progress towards reinforcement learning systems that generalize across diverse real-world tasks. Our code and experiments are available at \url{https://github.com/tidiane-camaret/contextual_rl_zero_shot}.

\end{abstract}

%%%%%%%%%%%%%%%%%%%%%%%%%%%%%%%%%%%%%%%%%%%%%%%%%%%%%%%%%%%%%%%%
%% Section: Submission of papers to RLC
%%%%%%%%%%%%%%%%%%%%%%%%%%%%%%%%%%%%%%%%%%%%%%%%%%%%%%%%%%%%%%%%

\section{Introduction}

Reinforcement Learning \citep[RL;][]{sutton-book18a} is a key area in creating intelligent systems that can learn and adapt to various tasks. A significant challenge in RL is crafting algorithms that perform well not only in known environments, but also in new, unseen ones. This highlights the need for generalizable RL agents. Generalization in RL includes a range of definitions and strategies, each suited to the specific requirements of different applications. The concept of generalization in RL covers everything from basic domain adjustments to advanced multitask learning settings \citep{kirk-jair23a}.

In this context, zero-shot generalization (ZSG) is a notable aspect of generalization. ZSG is unique because it requires that learned policies can be applied to new environments without additional modifications during testing \citep{kirk-jair23a,benjamins-tmlr23a}. This is especially important for applications where it is not possible to fine-tune the model in the target environment. ZSG is attractive because it suggests that RL models can effectively handle the diversity and unpredictability of real-world tasks. However, RL algorithms often struggle with even minor changes in their environment, which can result in behaviors that do not transfer well \citep{henderson-aaai18a,andrychowicz-iclr21a}. We attribute this issue primarily to the complexity of the RL training process, which typically does not focus on generalization \citep{parkerholder-jair22a}.

The path to achieving reliable ZSG for RL presents many challenges.
In particular, how can zero-shot generalization be achieved if we do not have access to privilidged information, i.e. a context \citep{hallak-arxiv15a,modi-alt18a}, that can help us identify the underlying transition dynamics.
For example, an RL agent assigned to control a robot may adjust its control behavior based on the weight of the load the robot needs to carry. If the robot is equipped with sensors that detect this weight, the RL policy could immediately use the information to adapt the control accordingly. However, if this information is not easily accessible, learning a policy that can adapt to such changes becomes even more difficult.
For this reason, recent works have proposed inferring a context from previous observations \citep[see, e.g.,][]{zhou_environment_2022,evans_context_2022}.
Although these works have shown great promise for zero-shot generalization, they often fall short of realizing their full potential.
We argue that this is largely due to the decoupling of learning the context from learning the policy. Instead, here, we propose learning behavior-specific context to aid policy learning by jointly learning context representations and well-performing policies.

In summary, this paper delves into the intricate domain of generalization in model-free RL, with a focus on zero-shot generalization from an inferred context.
Specifically, the contributions of our work are as follows:
\begin{itemize}\itemsep0pt
    \item We propose a novel RL algorithm, capable of zero-shot generalization by learning to recognize the environment dynamics while jointly learning desirable behaviors within the environment.
    \item Our empirical evaluation of the proposed algorithm contrasts the learning behavior of RL agents when providing explicit access to a ground truth context versus an inferred context.
    \item We provide insights into the learned context embeddings and show that they recover the relationship between transition dynamics.
\end{itemize}

\section{Related Work}

Various approaches have been proposed to learn generalizable RL agents. In this line of research, two particular areas have gained more momentum with a focus on either few-shot or zero-shot generalization abilities.

\paragraph{Meta-RL}
Meta-reinforcement learning \citep[Meta-RL;][]{beck-arxiv23a} involves the process of learning how to learn in reinforcement learning. Thus, the goal is to learn RL pipelines that can learn efficiently, so that they can be easily transferred to new settings and learn to solve a new task with few interactions. For example, \citet{duan-arxiv16a} proposed to encode the learning dynamics of a proximal policy optimization \citep[PPO;][]{schulman-arxiv17a} in a recurrent neural network (RNN). Thus, by giving examples of how the PPO agent adapts its behavior over time during training, the RNN can learn how to adapt a policy, without needing to perform gradient updates at test time. In an ideal scenario, such a learned RL agent only needs a few exploration episodes to find the optimal behavior, i.e., few-shot adaptation. To identify environment dynamics, most of such model free meta-RL agents \citep[see, e.g.,][]{finn-icml17a,wang-cogsci17a,rakelly_efficient_2019,nagabandi-iclr19a,melo-icml22a} and model-based ones \citep[see, e.g.,][]{lee-icml20b,guo-iclr22a,sodhani-ldcc22a,wen-arxiv23a} keep a short history of transitions to estimate the environment transition dynamics. Although much progress has been reported in meta-RL, at test time, many thousands of environment interactions are required for the learned RL agents to solve the test tasks reliably.

\paragraph{Contextual RL}
In contextual RL (cRL), it is assumed that the underlying transition dynamics can be characterized by a context \citep{hallak-arxiv15a,modi-alt18a}. This could, for example, be a physical property such as wind \citep{koppejan-gecco09a}, the length of the pole that needs to be balanced \citep{seo-neurips20a,kaddour-neurips20a,benjamins-tmlr23a}, terrain characteristic \citep{escontrela-arxiv20a}, or more abstract concepts that characterize the underlying environment dynamics \citep{biedenkapp-ecai20a,adriaensen-jair22a}. \citet{kirk-jair23a} identify the cRL setting as particularly relevant for the study of zero-shot generalization capabilities of RL agents, as the cRL framework allows to define the ranges of inter- and extrapolation distributions and enable a systematic and principled study of how RL agents can adapt to changes in their environments. Using the evaluation protocol proposed by \citet{kirk-jair23a}, \citet{benjamins-tmlr23a} studied the generalizability of multiple model-free RL agents on a benchmark that uses various physical properties as context information. Their study assumed a naive use of context information by concatenating it directly with the observation. Instead, \citet{beukman_dynamics_2023} proposed to use a hypernetwork to learn adaptable RL agents. However, their approach requires that agents can directly observe the context. In contrast, our work studies the effectiveness of inferring context, as typically done in meta-RL, for zero-shot generalization without assuming access to the context.

\section{Background - Contextual Markov Decision Processes}

We build upon the framework of Contextual Markov Decision Processes \citep[cMDPs;][]{hallak-arxiv15a,modi-alt18a} 
which was proposed as an ideal abstraction for the study of zero-shot generalization in RL \citep{kirk-jair23a}.
In an MDP $\MDP = (\statespace, \actionspace, \transdomain, \rewards, \rho)$, the characteristics of the environment are represented by the state space $\statespace$, the action space $\actionspace$, the transition dynamics $\transdomain$, the reward function $\rewards$, and the initial state distribution $\rho$. 

cMDPs introduce context to parameterize an environment's behavioral rules, allowing for variations in task instances. In a cMDP $\mathcal{M}_\text{cMDP}$, the action space $\mathcal{A}$ and the state space $\mathcal{S}$ remain untouched, while the transition dynamics $\mathcal{T}_\text{c}$, the reward $\mathcal{R}_\text{c}$ and the initial state distribution $\rho_\text{c}$ are context dependent and vary with context $\text{c} \in C$. The context-aware initial state distribution $\rho_\text{c}$ and dynamic changes expose the agent to different parts of the state space across contexts. Therefore, a cMDP $\mathcal{M}_\text{cMDP}$ encompasses a set of MDPs that vary by context, denoted as $\{\mathcal{M}_c \}_{\text{c} \in C}$, which provides a framework for studying generalization in RL in diverse environments.

In the context of zero-shot problems, we can then utilize cMDPs to study generalization by defining two sets of context sets $C_{train}$ and $C_{eval}$, for training and evaluation, respectively \citep{kirk-jair23a}. During training, the context values are sampled from $C_{train}$. In this set, agents can learn to use or infer the context before being subsequently evaluated with the values of $C_{eval}$. To evaluate true generalization capabilities, $C_{eval}$ needs to be disjoint from $C_{train}$, however, their ranges can overlap.
If all context values of $C_{eval}$ fall within the limits of $C_{train}$, this allows us to 
measure the \emph{interpolation} performance of the agent.
However, if they are all outside those limits, we can measure an agents \emph{extrapolation} performance.
Furthermore, if the context is higher dimensional, we can study a combination of both, where we assess the performance of extrapolation along one dimension, while studying the interpolation along the other \citep{kirk-jair23a}.

\citet{benjamins-tmlr23a} followed this protocol to study the generalizability of various model-free agents.
In particular, they contrasted providing direct access to the context (as part of the observation of the agent) with simply learning on a distribution (similar to domain randomization \citep{tobin-iros17a,peng-icra18a}) without having direct access to the context. Their findings showed that direct access to context is not always beneficial and how best to incorporate contextual information into an agent remains an open question.
However, in particular, \citet{benjamins-tmlr23a} did not contrast this with agents that learn to infer context.
In contrast, here, we study ZSG for agents that first need to infer their environment context and contrast this with both agents that have no access to the context and those that can directly observe it.

\section{Method}\label{sec:method}
We begin this section by discussing how past experiences can be used to infer the context of the environment. Further, we discuss the advantages and disadvantages of this style of inferring context in the few-shot and zero-shot settings, respectively, before using these insights to derive methodology that is particularly suited to the zero-shot setting. Finally, we outline the proposed learning algorithm that follows this methodology.

\subsection{Inferring Context From Past Experiences}\label{sec:infer}
\begin{figure}[ht]
    \centering
    \begin{subfigure}[b]{0.49\textwidth}
    \includegraphics[width=\textwidth]{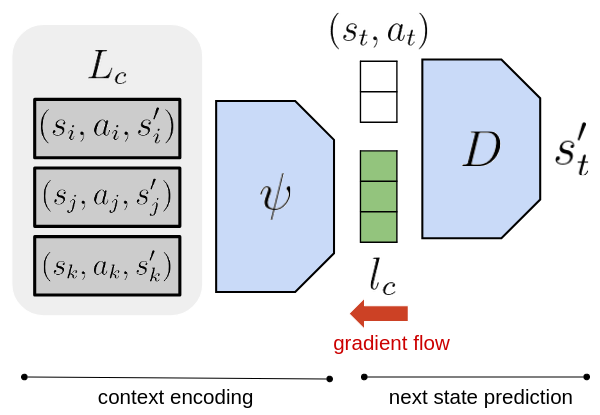}
    \caption{Training of the context encoder $\psi$}
    \label{fig:training_iida_1}
    \end{subfigure}
    \begin{subfigure}[b]{0.49\textwidth}
    \includegraphics[width=\textwidth]{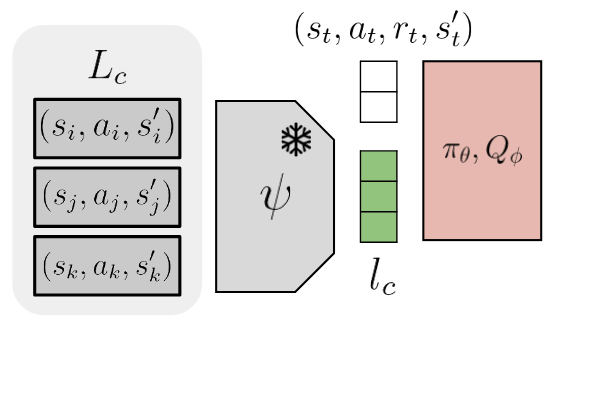}
    \caption{Training of the policy networks $\pi_\theta$, $Q_\phi$.}
    \label{fig:training_iida_2}
    \end{subfigure}
    \caption{Two-phase training of predictive context encoding methods. Typically, no gradient updates go through the frozen context encoder while learning the policy as depicted in (\protect\subref{fig:training_iida_2}).}
    \label{fig:2stage-dyn-pred}
\end{figure}
The task of learning to infer the dynamics of the environment is generally done by looking at past experience. Typically, this involves training a context encoder $\psi$ against an auxiliary training objective \citep[see, e.g.,][]{zhou_environment_2022,evans_context_2022}. Given a list $L$ of transitions collected in an environment, the context encoder $\psi$ aims to generate a latent representation $l$ that captures relevant contextual information. 
Predominantly, training of $\psi$ involves learning to predict the one-step dynamics of the environment from past trajectories based on the encoding learned in $\psi$. The goal of this style of learning context embedding is to directly capture how the environment evolves with each action taken.

Concretely, consider a list of observed transitions $L_c = [(s_j, a_j, s'_j) \mid j \in [0, h-1]]$ collected within a specific instance of the environment given context $c$, where $s_j$ represents the state, $a_j$ denotes the action, and $s'_j$ the subsequent state in transition $j$. The encoder $\psi$ uses these transitions to learn a lower-dimensional representation $l_c$. The latent representation is then used by the dynamics predictor $D$ together with the current state-action pair $(s,a)$ and tasked with predicting the next state $s'$. The prediction error of $D$ is then backpropagated through $\psi$ to ensure that $\psi$ captures the information relevant to predict the one-step dynamics. Schematically, this can be seen in \Cref{fig:training_iida_1}.
Once the context encoder $\psi$ is fully trained, the learned embedding is fixed and used to produce contextual information during subsequent training of the policy, see \Cref{fig:training_iida_2}.

Inferring dynamics from previous observations has been shown to be particularly useful for few-shot adaptation in meta-RL \citep[see, e.g.,][]{rakelly_efficient_2019,nagabandi-iclr19a,melo-icml22a}\footnote{For a recent survey on meta-RL, we refer to \citep{beukman_dynamics_2023}}. In this setting, the (learned) RL agent is allowed to update its own policy based on new observations (i.e., few shots) in the new environment. Thus, a context that aims to capture the entire transition dynamics will be helpful in exploring novel behaviors that might traverse drastically different parts of the state space. However, we argue that inferring context for zero-shot adaptation of model-free agents should not consider a learned context based on the full transition dynamics, but rather one that only captures that information, which is particularly relevant to the learned policy, which will remain unchanged at test time.

\subsection{The Case for Behavior-Specific Context for Zero-Shot Generalization}
\label{sec:infer_context_exp}
Consider the example of a policy that needs to control a 4-legged robot to navigate through different worlds with varying properties, such as friction levels or gravity.
At test time, in the few-shot setting, the robot might explore the landscape of the current world. Any observations made during these initial steps can be used to refine its behavior policy (e.g., change the gait to better suit the environment), and the task can be repeated as often as the particular few-shot setting allows.
However, in the zero-shot setting, at test time, there is only one chance to get it right. The policy needs to be spot on from the get-go, as there is no further fine-tuning or other refinement of the policy. Thus, the context serves very different purposes in both scenarios. 

The question now arises: How can we get a context that is maximally helpful to the policy at hand? Ideally, such a context should capture all the intricacies that the policy might encounter when interacting with the environment. Essentially, we want a context that is conditioned on the policy at hand. At first glance, this leads to a chicken-and-egg problem in which we need to learn a policy-conditioned context so that we can learn a context-conditioned policy or vice versa. Instead, to avoid this issue, we can use the general learning dynamics in RL that typically follows the principle of generalized policy iteration \citep[GPI;][]{sutton-book18a}.

\begin{wrapfigure}{r}{0.5\textwidth}\vskip-.1in

    \centering
    \includegraphics[width=0.49\textwidth]{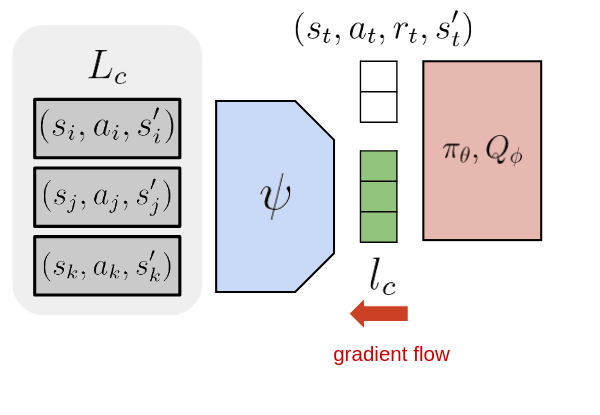}%
    \vskip-.3in
    \caption{Joint training of the context encoder $\psi$ and policy/value networks $\pi_\theta$/$Q_\phi$}%
    \label{fig:method}%
    \vskip-.1in

\end{wrapfigure}
In GPI we iterates between two stages (I) policy evaluation and (II) policy iteration. Starting from a random policy, we evaluate its performance. Using the evaluation data, we then improve our policy, etc. Thus, as we traverse the policy space, we get many samples of different behavior policies. These samples can be used directly to learn the latent context representation from observed past experiences in a style similar to that described in \Cref{sec:infer}. To condition the context encoder on the policy behavior, we backpropagate losses of the policy objectiv through the context encoder. Thus, at test time, our context will capture all the information that is relevant to the current behavior of the policy. \Cref{fig:method} outlines our proposed joint context and behavior learning scheme.

\begin{algorithm}[hb]
    \caption{Training loop}\label{alg:training_loop}
    \begin{algorithmic}[1]
    \State Initial state $s_0$, context value $c$, context encoder $\psi$, actor and critic networks $\pi_\theta$, $Q_\phi$, number of training steps $T_{train}$, transition list size $h$, batch size $b$, replay buffer $R = \emptyset$
    \For {$t \in [0,T_{train} - 1 ]$}:
        \State
        \State \textbf{Action Sampling}\label{algo:sampling-start}
        \State $L_c \gets Sample( \{ (s_j, a_j, s'_j) \mid (s_j, a_j, r_j, s'_j, c_j) \in R, c_j = c \}, size = h)$ \Comment{Sample a list $L_c$ of transitions from context $c$}
        \If{$L_c \neq \emptyset$}
            \State $l_c \gets \psi(L_c)$ \Comment{Infer the context latent $l_c$}
        \Else
            \State $l_c \gets 0$
        \EndIf
        \State $a_t \sim \pi_\theta(\cdot|(s_t,l_c))$ \Comment{Sample action $a_t$ from policy $\pi_\theta$ based on the augmented state $(s_t,l_c)$ }
        \State $s'_t \sim \mathcal{T}_\text{c}(\cdot|(s_t,a_t)) , r_t \sim \mathcal{R}_\text{c}(\cdot|(s_t,a_t))$ \Comment{Execute action $a_t$ and observe next state $s'_t$, reward $r_t$}
        \State $R \gets R \cup (s_t, a_t, r_t, s'_t,c)$ \Comment{Store transition in replay buffer}\label{algo:sampling-end}
        \State
        \State \textbf{Policy update}
        \State $B \gets Sample(R, size = b)$ \Comment{Sample a mini-batch of transitions from the replay buffer}\label{algo:update-start}
        \For {$(s_i, a_i, r_i, s'_i, c_i) \in B$}:
            \State $L_{c_i} \gets Sample( \{(s_j, a_j, s'_j) \mid\ (s_j, a_j, r_j, s'_j, c_j) \in R, c_j = c_i \})$ \Comment{Sample a list $L_{c_i}$ of transitions from context $c_i$}
            \State $l_{c_i}\gets \psi(L_{c_i})$ \Comment{Infer the context latent $l_{c_i}$}
        \EndFor
        \State Compute the loss of the actor and critic networks using the mini-batch of augmented transitions $\{((s_i,l_{c_i}), a_i, r_i, (s'_i,l_{c_i})) \mid (s_i, a_i, r_i, s'_i, c_i) \in B \}$
        \State Update the context encoder $\psi$, actor $\pi_\theta$ and critic $Q_\phi$ networks\label{algo:update-end}
    \EndFor
    \end{algorithmic}
\end{algorithm}
\subsection{Joint Context and Policy Learning}
    We present our proposed learning approach \Cref{alg:training_loop}. The pseudocode is written from the perspective of actor-critic style learning, such as the soft actor critic algorithm \citep[SAC;][]{haarnoja-icml18a}, an off-policy deep reinforcement learning algorithm that simultaneously learns a stochastic policy $\pi_\theta$ and a state-action-value function $Q_\phi$. However, it is important to note that our method is adaptable and can be integrated with any off-policy algorithm.

    Lines \ref{algo:sampling-start} to \ref{algo:sampling-end} detail how an agent can infer its context and act with respect to it. We first sample a list of $h$ past transitions to predict the latent context $l_c = \psi(L_c)$. We then append this learned context to the current state observation $s_t$, allowing us to condition the policy on both state and context.
    When training the policy (lines \ref{algo:update-start} to \ref{algo:update-end}), we first infer the context from past transitions in a similar way. When updating the context encoder $\phi$, we propagate the error through the actor to condition the context on the policy.

 During training, transitions are uniformly sampled from the entire replay buffer in order to ensure diverse experiences for the context encoder. However, at inference time, we assume that the agent only has access to the transitions of the current episode, see \Cref{alg:eval_loop}.

\begin{algorithm}[t]
    \caption{Evaluation loop}\label{alg:eval_loop}
    \begin{algorithmic}[1]
    
    \State Initial state $s_0$, context encoder $\psi$, actor and critic networks $\pi_\theta$, $Q_\phi$,  number of episode steps $T_{episode}$, transition list $L = \emptyset$, latent context $l = 0$
    \For {$t \in [0,T_{episode} - 1]$}:
        
        \If{$L \neq \emptyset$}
            \State $l \gets \psi(L)$ \Comment{Infer the context latent $l$}
        \EndIf
        \State $a_t \sim \pi_\theta(\cdot|(s_t,l))$ \Comment{Sample action $a_t$ from policy $\pi_\theta$ based on the augmented state $(s_t,l)$ }
        \State $s_{t+1} \sim \mathcal{T}_\text{c}(\cdot|(s_t,a_t))$ \Comment{Execute action $a_t$ and observe next state $s'_t$}
        \State $L \gets L \cup (s_t, a_t, s'_t)$ \Comment{Store transition $(s_t, a_t, s'_t)$ in $L$}
    \EndFor
    \end{algorithmic}
\end{algorithm}

\section{Experiments}

\subsection{Environments}
We evaluate our context encoding method across multiple environments as described below, using the CARL library \citep{benjamins-tmlr23a}, which allows adjustments to the dynamics parameters during and between episodes. Here, we provide a brief description of the environments and contexts. For the exact training and evaluation context sets, see \Cref{tab:environment_context} in the Appendix.
\begin{itemize}\itemsep0em
    \item \textbf{Cartpole:} An agent needs to learn to balance a pole vertically on a moving cart which it controls. The observation space is a 4-dimensional vector (position, velocity of the cart, angle, angular velocity of the pole). We evaluate generalizability to changes in time elapsed between states.
    \item \textbf{Pendulum:} An agent has to learn to swing up an inverted pendulum and stabilize it at the top from a random initial position. The action space controls force direction and magnitude. We assess robustness to changes in the pendulum's length.
    \item\textbf{MountainCar:} The agent has to learn to drive a car up a steep slope, potentially needing to gain momentum using the opposite slope. We test adaptability to changes in power, both positive and negative values are tested. In our setting, the car is given various levels of power, allowing much faster goal reaches, potentially allowing for positive rewards.
    \item \textbf{Ant:} The agent has to learn to control a 4-legged robot to facilitate walking. The observation space includes joint angles, velocities, contact forces, and torques (27 dimensions). We evaluate against changes in the robots torso mass.
\end{itemize}

\subsection{Baseline methods}
To assess the generalization ability of our method, we compare it against the following baselines:
\begin{itemize}\itemsep0em
    \item \textbf{Hidden context:} No context is explicitly provided to the agent, but it is still trained across all contexts in the training set. This can be seen as similar to domain randomization \citep{tobin-iros17a}. This provides a \emph{lower bound} for generalization performance, as we constrain the agent to learn a single, non-adaptive policy.
    
    \item \textbf{Explicit identification:} The explicit context value $c$ is concatenated to the observed state at each time step, at training and evaluation time.  In the contextual RL setting, this is the most widely used way of learning with access to context information \citep{benjamins-tmlr23a}. As the policy has access to complete information about the environment dynamics, we consider this setting as an \emph{upper bound} for generalization performance, as agents are provided with the ground truth context.
    
    \item \textbf{Predictive identification:} In this method, a context encoder is trained on the transition dynamics prediction task described in \Cref{sec:infer_context_exp}, following the training pipeline from \cite{evans_context_2022}. This baseline allows us to evaluate the effectiveness of jointly learning context and policy compared to learning them separately.
\end{itemize}
All baseline learning methods and our proposed method for learning with and from context are evaluated with a soft actor-critic \citep[SAC;][]{haarnoja-icml18a} agent. We keep the SAC hyperparameters fixed for all methods (see \Cref{tab:sac_hyperparameters} in the Appendix). We report the hyperparameters relevant for learning to infer contexts in \Cref{tab:encoder_hyperparameters} in the Appendix.
    
\subsection{Measures of generalization}
For each environment and method, we perform $10$ independent training runs (using different random seeds) and then evaluate the trained agent for $20$ episodes, resulting in a total number of $N = 200$ evaluations for each value of $C_{eval}$. This produces a score matrix $(s_{c_i,n})$, with $c_i \in C_{eval}$ and $n \in [1,N]$, per environment and per method. To compare scores between different context values, we follow the methodology described by \citet{agarwal-neurips21a} and normalize the scores by linearly rescaling them based on two reference points. We note the score of a random agent as $s^{random}_{c_i}$ and the score of an agent trained only on the default value of the context as $s^{default}_{c_i}$, and calculate the normalized scores $\Bar{s}_{c_i,n}$ as:
$$\Bar{s}_{c_i,n} = \frac{s_{c_i,n} - s^{random}_{c_i}}{s^{default}_{c_i} - s^{random}_{c_i}}$$

We compute the \textbf{interquartile mean (IQM)} of the agent's performance on both the interpolation and extrapolation subsets of $C_{eval}$, along with their respective stratified bootstrap confidence intervals, as outlined by \citet{colas2019hitchhikers}.

\begin{table}[tbp]
    \centering
    \begin{tabular}{l|r|r|r|r}
        \toprule
        \multirow{6}{*}{Environment} & \multicolumn{4}{c}{Episodic return - area under the  curve (average over 10 seeds)} \\
        \cmidrule{2-5}
        & \multirow{3}{*}{Hidden context} & \multirow{3}{*}{Explicit context} & \multirow{1}{*}{Predictive} & Joint context \&\\
        &&&\multirow{1}{*}{identification}&policy learning\\
        \midrule
        Cartpole & $\mathbf{287\,170}$ & $283\,847$ & $285\,018$ & $285\,528$ \\
        Pendulum & $-115\,649$ & $-80\,745$ & $-108\,771$ & $\mathbf{-78\,892}$ \\
        MountainCar & $1\,722$ & $\mathbf{1\,104\,599}$ & $103\,512$ & $510\,677$\\
        Ant & $57\,552$ & $-20\,673$ &  $53\,651$ & $\mathbf{62\,282}$ \\
        \bottomrule
    \end{tabular}
    \caption{Learning progression across \emph{training} environments expressed as the area under the reward curve. Our method consistently achieves high AUC for all environments.}
    \label{tab:auc_scores}
\end{table}
\begin{figure}[tbp]
    \centering
    \begin{subfigure}[b]{0.49\textwidth}
    \includegraphics[width=\textwidth]{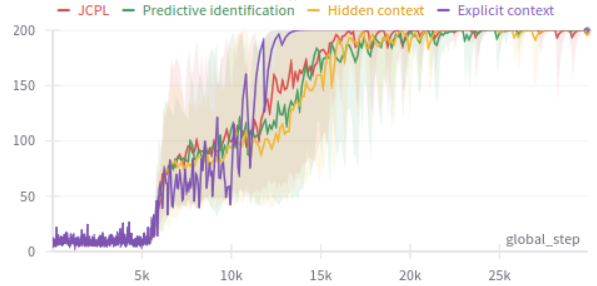}
    \caption{Cartpole}
    \label{fig:learning_curve_cartpole}
    \end{subfigure}
    \begin{subfigure}[b]{0.49\textwidth}
    \includegraphics[width=\textwidth]{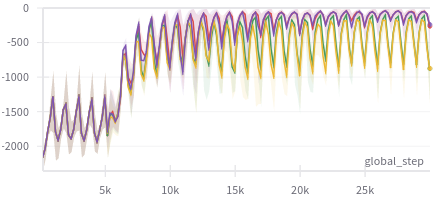}
    \caption{Pendulum}
    \label{fig:learning_curve_pendulum}
    \end{subfigure}
    \\
    \begin{subfigure}[b]{0.49\textwidth}
    \includegraphics[width=\textwidth]{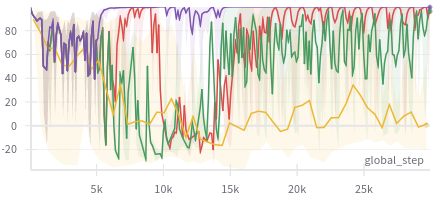}
    \caption{MountainCar}
    \label{fig:learning_curve_mountaincar}
    \end{subfigure}
    \begin{subfigure}[b]{0.49\textwidth}
    \includegraphics[width=\textwidth]{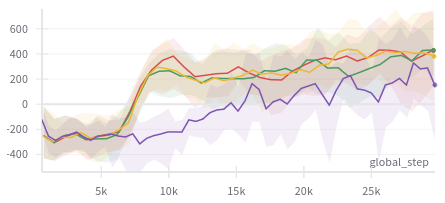}
    \caption{Ant}
    \label{fig:learning_curve_ant}
    \end{subfigure}
    \caption{Episodic returns during training when learning with explicit context, hidden context and learned context embeddings. Our joint learning method (jcpl) is capable of reaching the same performance as if learning with the ground-truth context, outperforming the predictive identification baseline.}
    \label{fig:learning_curves}
\end{figure}
\subsection{Research questions}
\paragraph{Research Question 1:} Is it advantageous to learn behavior-specific context embeddings?

During training, our method consistently exhibits higher episodic returns compared to the predictive identification method. We display the learning curves in Figure \ref{fig:learning_curves}, as well as the area under the training curve in Table \ref{tab:auc_scores}. On the Pendulum and Ant tasks, our proposed method of joint context and policy learning achieves the highest episodic returns, outperforming every baseline. On the 4 tasks, it outperforms the predictive context identification method. However, in the Cartpole environment, learning a single policy without explicit context can also perform competitively.

\begin{figure}[tbp!]
    \centering
    \begin{subfigure}[b]{0.49\textwidth}
    \includegraphics[width=\textwidth]{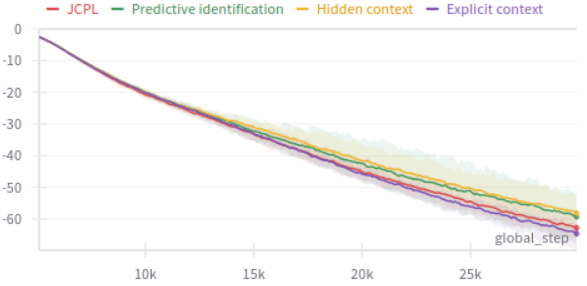}
    \caption{Cartpole - actor loss}
    \label{fig:actor_loss_cartpole}
    \end{subfigure}
    \begin{subfigure}[b]{0.49\textwidth}
    \includegraphics[width=\textwidth]{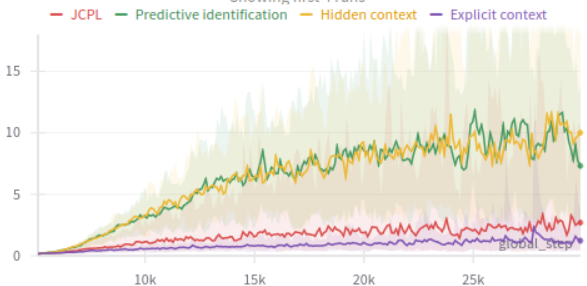}
    \caption{Cartpole - critic loss}
    \label{fig:critic_loss_cartpole}
    \end{subfigure}
    \\
    \begin{subfigure}[b]{0.49\textwidth}
    \includegraphics[width=\textwidth]{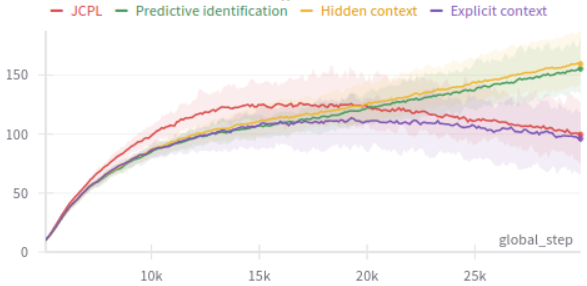}
    \caption{Pendulum - actor loss}
    \label{fig:actor_loss_pendulum}
    \end{subfigure}
    \begin{subfigure}[b]{0.49\textwidth}
    \includegraphics[width=\textwidth]{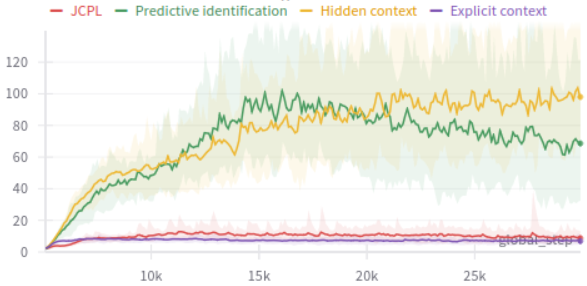}
    \caption{Pendulum - critic loss}
    \label{fig:critic_loss_pendulum}
    \end{subfigure}
    \caption{Losses of the policy model training when learning with explicit context, hidden context and learned context embeddings. Loss is consistently lower when jointly learning the context representation and the task policy.}
    \label{fig:loss-curves}
\end{figure}

We also note that the loss curves of the actor and critic models are consistently lower when the context and task policy are jointly learned, as shown in Figure \ref{fig:loss-curves}. In particular, the loss curves for Cartpole and Pendulum indicate that our approach enables the agent to leverage contextual information as effectively as having direct access to the true context variable. By tailoring the learned context representations to the specific needs of the policy, our method can capture the relevant environmental dynamics that facilitate better generalization across different contexts during evaluation.

\paragraph{Research Question 2:} Are behavior-specific context embeddings beneficial for zero-shot out-of-distribution generalization?

\begin{figure}[tb]\vskip-.2in
    \centering
     \begin{subfigure}[b]{0.40\textwidth}
    \includegraphics[width=\textwidth]{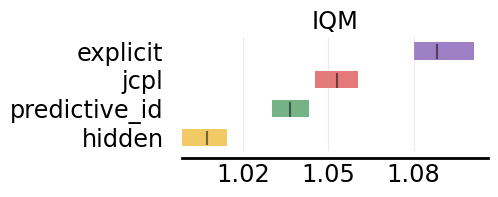}
    \caption{Cartpole - interpolation}
    \label{fig:exp-comparison-CI}
    \end{subfigure}
    \begin{subfigure}[b]{0.40\textwidth}
    \includegraphics[width=\textwidth]{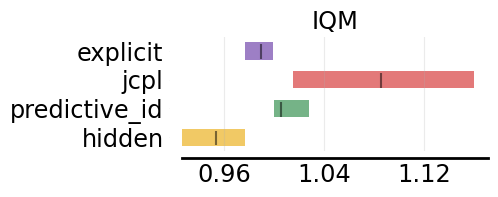}
    \caption{Cartpole - extrapolation}
    \label{fig:exp-comparison-CE}
    \end{subfigure}
    \\
    \begin{subfigure}[b]{0.40\textwidth}
    \includegraphics[width=\textwidth]{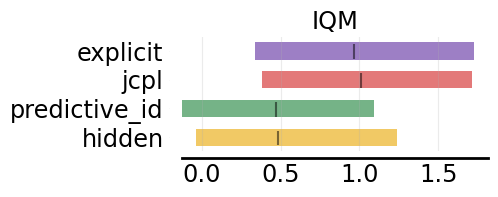}
    \caption{Pendulum - interpolation}
    \label{fig:exp-comparison-PI}
    \end{subfigure}
    \begin{subfigure}[b]{0.40\textwidth}
    \includegraphics[width=\textwidth]{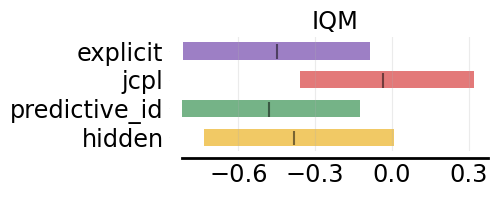}
    \caption{Pendulum - extrapolation}
    \label{fig:exp-comparison-PE}
    \end{subfigure}
    \\
    \begin{subfigure}[b]{0.40\textwidth}
    \includegraphics[width=\textwidth]{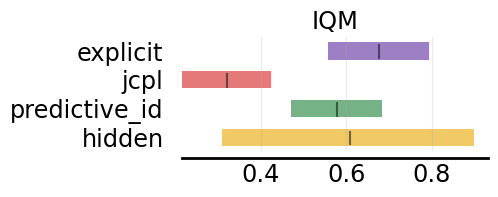}
    \caption{MountainCar - interpolation}
    \label{fig:exp-comparison-MI}
    \end{subfigure}
    \begin{subfigure}[b]{0.40\textwidth}
    \includegraphics[width=\textwidth]{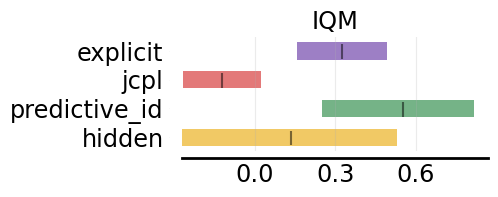}
    \caption{MountainCar - extrapolation}
    \label{fig:exp-comparison-ME}
    \end{subfigure}
    \\
    \begin{subfigure}[b]{0.40\textwidth}\includegraphics[width=\textwidth]{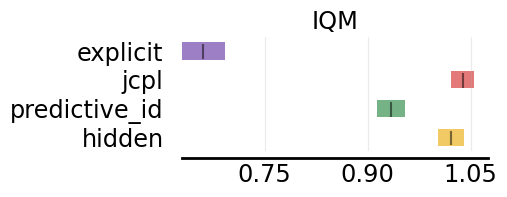}
    \caption{Ant - interpolation}
    \label{fig:exp-comparison-AI}
    \end{subfigure}%
    \begin{subfigure}[b]{0.40\textwidth}\includegraphics[width=\textwidth]{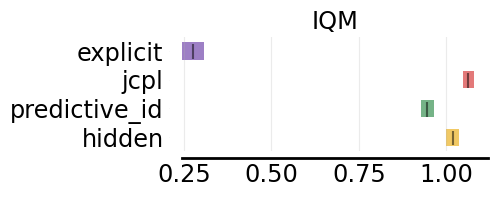}
    \caption{Ant - extrapolation}
    \label{fig:exp-comparison-AE}
    \end{subfigure}
    \caption{Interquartile Mean (IQM) of the aggregated normalized scores, along with their respective stratified bootstrap confidence intervals, in the interpolation and extrapolation settings}\vskip-.2in
    \label{fig:generalization-metrics-extra}
\end{figure}

We evaluate the trained agents on the $C_{eval}$ sets, and display the IQMs of each method \Cref{tab:generalization_metrics}. For an idea of the distribution of the scores, we also show the confidence intervals of the IQMs in \Cref{fig:generalization-metrics-extra}. In Cartpole and Pendulum environments, our joint context and policy learning (jcpl) method achieves IQM values compared to the predictive identification baseline across all settings, including interpolation, extrapolation, and considering all context values. 

In particular, for the Ant environment, which is relatively more complex, the results strongly favor our jcpl method, with significantly higher IQM values compared to the predictive identification baseline. Interestingly, in this environment, the explicit context baseline, which directly provides the context value as input, leads to substantially worse performance than both jcpl and the predictive identification method. This corroborates the findings of \citet{benjamins-tmlr23a}: direct access to context is not always beneficial, and how best to incorporate contextual information into an agent remains an open question. Our approach of jointly learning behavior-specific context embeddings directly addresses this open question and demonstrates improved generalization performance, especially in complex environments like Ant.

To provide further clarity for the reader, we depict the full empirical distribution of the score of each method in the Appendix (\Cref{fig:generalization-metrics-intra-extra,fig:generalization-metrics-all}), in the form of \textbf{performance profiles}.

\begin{table}[tbp]
    \centering
    \begin{tabular}{l|l|c|c|c}
    %\hline
    \toprule
    Environment & Metric & Interpolation & Extrapolation  & All values \\\midrule

    \multirow{2}{*}{Cartpole} & IQM jcpl & \textbf{1.052939} & \textbf{1.085407} & \textbf{1.029779}   \\ 
    \cmidrule{2-5}
    & IQM predictive id & 1.036612 & 1.005375 & 1.017251\\ 
    %\cmidrule{2-5}
    %& Probability of improvement & 0.544375 & 0.510833 & 0.519400\\ 
    
    \midrule  
    \multirow{2}{*}{Pendulum} & IQM jcpl & \textbf{1.012740} & \textbf{-0.034381} & \textbf{0.384468}   \\ 
    \cmidrule{2-5}
    & IQM predictive id & 0.467899 & -0.478549 & -0.104004 \\ 
    %\cmidrule{2-5}
    %& Probability of improvement & 0.610938 & 0.515139 & 0.535100 \\ 
    \midrule    
    \multirow{2}{*}{MountainCar} & IQM jcpl & 0.320692 & -0.123045 & 0.091731   \\ 
    \cmidrule{2-5}
    & IQM predictive id & \textbf{0.578874} & \textbf{0.553004} & \textbf{0.588501} \\ 
    %\cmidrule{2-5}
    %& Probability of improvement & 0.425000 & 0.412639 & 0.429545 \\ 
    \midrule    
    \multirow{2}{*}{Ant} & IQM jcpl & \textbf{1.038206} & \textbf{1.063549} & \textbf{1.051768}   \\ 
    \cmidrule{2-5}
    & IQM predictive id & 0.934375 & 0.946111 & 0.940794 \\ 
    %\cmidrule{2-5}
    %& Probability of improvement & 0.562800 & 0.571250 & 0.565207\\ 

    %\hline
    \bottomrule
    \end{tabular}
    \caption{Generalization metrics for the normalized scores of the jcpl and predictive identification methods. In every environmnent except MountainCar, our method scores a higher IQM, both in the Interpolation and extrapolation settings.}
    \label{tab:generalization_metrics}
\end{table}

\paragraph{Research Question 3:} Do the learned embeddings capture the underlying ground truth change in the transition dynamics?

\begin{figure}[tbp]
    \centering
    \begin{subfigure}[b]{.24\textwidth}
    \includegraphics[width=\textwidth]{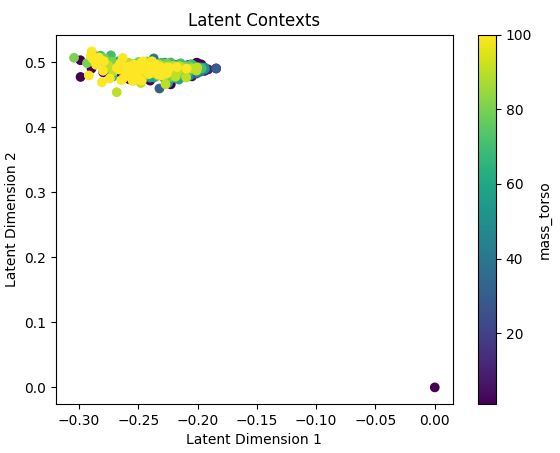}
    \caption{Ant\\predictive identification}
    \label{fig:exp-embedding-env-a-s-1}
    \end{subfigure}
    \begin{subfigure}[b]{.24\textwidth}
    \includegraphics[width=\textwidth]{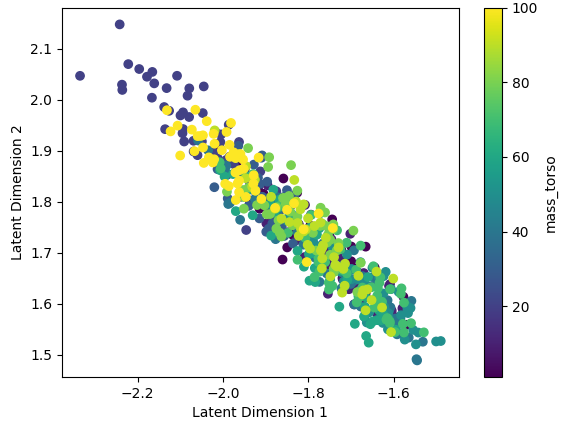}
    \caption{Ant\\jcpl}
    \label{fig:exp-embedding-env-a-s-2}
    \end{subfigure}
    \begin{subfigure}[b]{.24\textwidth}
    \includegraphics[width=\textwidth]{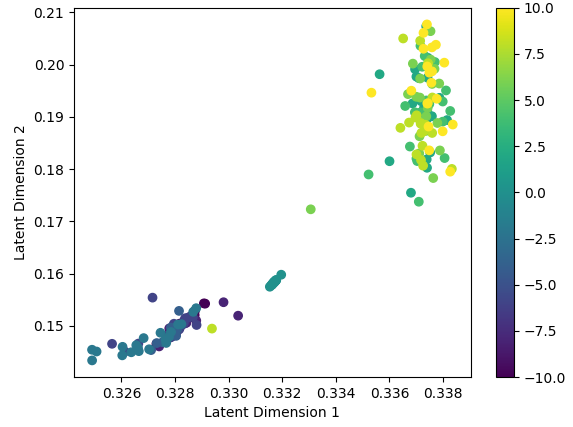}
    \caption{MountainCar\\predictive identification}
    \label{fig:exp-embedding-env-b-s-1}
    \end{subfigure}
    \begin{subfigure}[b]{.24\textwidth}
    \includegraphics[width=\textwidth]{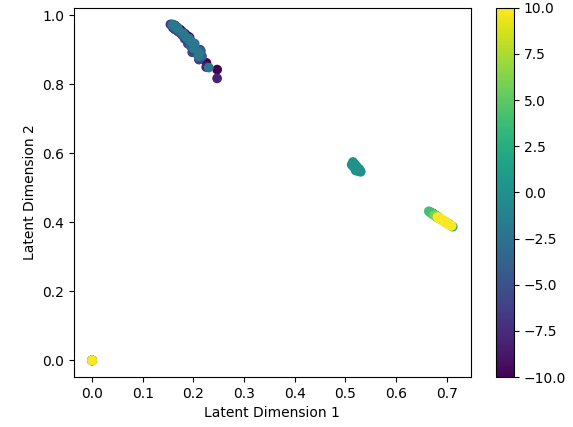}
    \caption{MountainCar\\jcpl}
    \label{fig:exp-embedding-env-b-s-2}
    \end{subfigure}
    \caption{Learned embeddings of both context encoding methods.}
    \label{fig:latents}
\end{figure}

To evaluate whether the learned context embeddings effectively capture the true underlying changes in the dynamics of the environment, we examine the 2D latent representations visualized in Figure \ref{fig:latents}. In Ant and MountainCar environments, the latent embeddings learned by our jcpl method exhibit better separation between different context values compared to the predictive identification baseline. This separation suggests that the jcpl embeddings encode information about the varying transition dynamics more distinctively. Further visualizations in the Appendix, across multiple environments and training seeds, reinforce this observation.

Quantitatively, we measure the mean squared error (MSE) of a random forest model when predicting the context value from the learned latent embeddings, using 5-fold cross-validation averaged over 10 training seeds (\Cref{tab:context_prediction}). In the Cartpole, MountainCar, and Ant environments, jcpl achieves a lower MSE compared to the predictive identification method, indicating that the learned embeddings better represent the ground-truth context information. Specifically, jcpl reduces MSE from $24.89$ to $15.86$ in MountainCar and from $1221.84$ to $1107.84$ in the complex Ant environment.

However, in the Pendulum environment, the predictive identification method achieves a slightly lower MSE of $0.0006$ compared to jcpl's MSE of $0.0008$. This suggests that learning the context separately may be more effective in certain environments for capturing the transition dynamics.

Overall, the qualitative and quantitative analysis demonstrate that our joint context and policy learning approach generally leads to learned embeddings that better capture the underlying ground-truth changes in the environment dynamics, particularly in more complex environments like Ant and MountainCar. By jointly optimizing the context and policy representations, jcpl can discover latent embeddings that encode relevant information about the varying transition dynamics, facilitating improved generalization.

\begin{table}[bp]
    \centering
    \begin{tabular}{l|p{2.5cm}|p{2.5cm}}
        %\hline
        \toprule
        \multirow{6}{*}{Environment} & \multicolumn{2}{|c}{Mean square error} \\
        \cmidrule{2-3}
        & Predictive identification & Joint context \& policy learning\\
        \midrule
        Cartpole &  0.0013 & \textbf{0.0003}  \\
        Pendulum & \textbf{0.0006} & 0.0008\\
        MountainCar & 24.89 & \textbf{15.86}\\
        Ant & 1221.84 & \textbf{1107.84} \\
        %\hline
        \bottomrule
    \end{tabular}
    \caption{Mean square error of random forest prediction of $c$ from $l_c$ (5-fold cross validation, mean value across 10 training seeds)}
    \label{tab:context_prediction}
\end{table}

\section{Conclusion}

In this paper, we introduce a novel approach that seamlessly integrates the learning of context-aware policies with the ability to infer contextual embeddings, addressing the critical challenge of zero-shot generalization in reinforcement learning. By enabling the joint optimization of policy and context representations, our algorithm acquires behavior-specific embeddings that significantly enhance the adaptability of RL systems to diverse environments, eliminating the need for retraining. This unified framework represents a substantial step towards creating more autonomous and versatile RL agents.

It is noteworthy that the experiments primarily focused on generalization across varying context values, without explicitly considering other factors such as task variations or changes in the reward structure. Consequently, the generalization capabilities of the method in scenarios involving such variations remain an open question that warrants further investigation. Future research should focus on improving the proposed method's generalization to varied tasks by incorporating reward signals into context modeling. This can be achieved by capturing task-specific information within the learned embeddings, enabling adaptation across diverse tasks. Observing reward signals during evaluation can further enhance context identification and generalization performance, expanding the method's applicability to a wider range of tasks with variations.

The current context encoder architecture averages latent context across transitions to produce a single embedding. However, exploring more advanced architectures that capture the evolution of transitions over time could yield richer and more informative context representations. By assessing the agent's uncertainty about its environment assessment, these architectures could enhance the agent's adaptability and performance across diverse environments, advancing context-aware reinforcement learning and enabling more robust and versatile autonomous agents in dynamic, real-world scenarios.

%\subsubsection*{Broader Impact Statement}
%\label{sec:broaderImpact}
%In this optional section, RLC encourages authors to discuss possible repercussions of their work, notably any potential negative impact that a user of this research should be aware of. 

\subsubsection*{Acknowledgments}
\label{sec:ack}
The authors acknowledge funding from The Carl Zeiss Foundation through the research network ``Responsive and Scalable Learning for Robots Assisting Humans'' (ReScaLe) of the University of Freiburg.
%Use unnumbered third level headings for the acknowledgments. All acknowledgments, including those to funding agencies, go at the end of the paper. Only add this information once your submission is accepted and deanonymized. 

%%%%%%%%%%%%%%%%%%%%%%%%%%%%%%%%%%%%%%%%%%%%%%%%%%%%%%%%%%%%%%%%
%% Bibliography
%%%%%%%%%%%%%%%%%%%%%%%%%%%%%%%%%%%%%%%%%%%%%%%%%%%%%%%%%%%%%%%%
\bibliography{bib/strings,bib/lib,bib/local,bib/proc}

\begin{thebibliography}{34}
\providecommand{\natexlab}[1]{#1}
\providecommand{\url}[1]{\texttt{#1}}
\expandafter\ifx\csname urlstyle\endcsname\relax
  \providecommand{\doi}[1]{doi: #1}\else
  \providecommand{\doi}{doi: \begingroup \urlstyle{rm}\Url}\fi

\bibitem[Adriaensen et~al.(2022)Adriaensen, Biedenkapp, Shala, Awad, Eimer, Lindauer, and Hutter]{adriaensen-jair22a}
S.~Adriaensen, A.~Biedenkapp, G.~Shala, N.~Awad, T.~Eimer, M.~Lindauer, and F.~Hutter.
\newblock Automated dynamic algorithm configuration.
\newblock \emph{Journal of Artificial Intelligence Research (JAIR)}, 75:\penalty0 1633--1699, 2022.

\bibitem[Agarwal et~al.(2021)Agarwal, Schwarzer, Castro, Courville, and Bellemare]{agarwal-neurips21a}
R.~Agarwal, M.~Schwarzer, P.~Samuel Castro, A.~C. Courville, and M.~G. Bellemare.
\newblock Deep reinforcement learning at the edge of the statistical precipice.
\newblock In M.~Ranzato, A.~Beygelzimer, K.~Nguyen, P.~Liang, J.~Vaughan, and Y.~Dauphin (eds.), \emph{Proceedings of the 35th International Conference on Advances in Neural Information Processing Systems ({N}eur{IPS}'21)}. Curran Associates, 2021.

\bibitem[Andrychowicz et~al.(2021)Andrychowicz, Raichuk, Sta{\'n}czyk, Orsini, Girgin, Marinier, Hussenot, Geist, Pietquin, Michalski, Gelly, and Bachem]{andrychowicz-iclr21a}
M.~Andrychowicz, A.~Raichuk, P.~Sta{\'n}czyk, M.~Orsini, S.~Girgin, R.{\"e}l Marinier, L.~Hussenot, M.~Geist, O.~Pietquin, M.~Michalski, S.~Gelly, and O.~Bachem.
\newblock What matters for on-policy deep actor-critic methods? a large-scale study.
\newblock In \emph{Proceedings of the International Conference on Learning Representations ({ICLR}'21)}, 2021.
\newblock Published online: \url{iclr.cc}.

\bibitem[Beck et~al.(2023)Beck, Vuorio, Liu, Xiong, Zintgraf, Finn, and Whiteson]{beck-arxiv23a}
J.~Beck, R.~Vuorio, E.~Z. Liu, Z.~Xiong, L.~Zintgraf, C.~Finn, and S.~Whiteson.
\newblock A survey of meta-reinforcement learning.
\newblock \emph{arXiv:2301.08028 [cs.LG]}, 2023.

\bibitem[Benjamins et~al.(2023)Benjamins, Eimer, Schubert, Mohan, Döhler, Biedenkapp, Rosenhan, Hutter, and Lindauer]{benjamins-tmlr23a}
C.~Benjamins, T.~Eimer, F.~Schubert, A.~Mohan, S.~Döhler, A.~Biedenkapp, B.~Rosenhan, F.~Hutter, and M.~Lindauer.
\newblock Contextualize me – the case for context in reinforcement learning.
\newblock \emph{Transactions on Machine Learning Research}, 2023.

\bibitem[Beukman et~al.(2023)Beukman, Jarvis, Klein, James, and Rosman]{beukman_dynamics_2023}
M.~Beukman, D.~Jarvis, R.~Klein, S.~James, and B.~Rosman.
\newblock Dynamics generalisation in reinforcement learning via adaptive context-aware policies.
\newblock In \emph{Proceedings of the 37th International Conference on Advances in Neural Information Processing Systems ({N}eur{IPS}'23)}. Curran Associates, 2023.

\bibitem[Biedenkapp et~al.(2020)Biedenkapp, Bozkurt, Eimer, Hutter, and Lindauer]{biedenkapp-ecai20a}
A.~Biedenkapp, H.~F. Bozkurt, T.~Eimer, F.~Hutter, and M.~Lindauer.
\newblock Dynamic algorithm configuration: Foundation of a new meta-algorithmic framework.
\newblock In J.~Lang, G.~De Giacomo, B.~Dilkina, and M.~Milano (eds.), \emph{Proceedings of the Twenty-fourth European Conference on Artificial Intelligence ({ECAI}'20)}, pp.\  427--434, June 2020.

\bibitem[Colas et~al.(2019)Colas, Sigaud, and Oudeyer]{colas2019hitchhikers}
C.~Colas, O.~Sigaud, and P.~Oudeyer.
\newblock A hitchhiker's guide to statistical comparisons of reinforcement learning algorithms.
\newblock \emph{RML@ICLR}, 2019.

\bibitem[Duan et~al.(2016)Duan, Schulman, Chen, Bartlett, Sutskever, and Abbeel]{duan-arxiv16a}
Y.~Duan, J.~Schulman, X.~Chen, P.~Bartlett, I.~Sutskever, and P.~Abbeel.
\newblock {RL{\textdollar}{\^{}}2{\textdollar}}: Fast reinforcement learning via slow reinforcement learning.
\newblock \emph{arXiv:1611.02779 [cs.AI]}, 2016.

\bibitem[Escontrela et~al.(2020)Escontrela, Yu, Xu, Iscen, and Tan]{escontrela-arxiv20a}
A.~Escontrela, G.~Yu, P.~Xu, A.~Iscen, and J.~Tan.
\newblock Zero-shot terrain generalization for visual locomotion policies.
\newblock \emph{arXiv:2011.05513 [cs.RO]}, 2020.

\bibitem[Evans et~al.(2022)Evans, Thankaraj, and Pinto]{evans_context_2022}
B.~Evans, A.~Thankaraj, and L.~Pinto.
\newblock Context is everything: Implicit identification for dynamics adaptation.
\newblock In \emph{International Conference on Robotics and Automation, ({ICRA}'22)}, pp.\  2642--2648. {IEEE}, 2022.

\bibitem[Finn et~al.(2017)Finn, Abbeel, and Levine]{finn-icml17a}
C.~Finn, P.~Abbeel, and S.~Levine.
\newblock Model-agnostic meta-learning for fast adaptation of deep networks.
\newblock In D.~Precup and Y.~Teh (eds.), \emph{Proceedings of the 34th International Conference on Machine Learning ({ICML}'17)}, volume~70, pp.\  1126--1135. Proceedings of Machine Learning Research, 2017.

\bibitem[Guo et~al.(2022)Guo, Gong, and Tao]{guo-iclr22a}
J.~Guo, M.~Gong, and D.~Tao.
\newblock A relational intervention approach for unsupervised dynamics generalization in model-based reinforcement learning.
\newblock In \emph{Proceedings of the International Conference on Learning Representations ({ICLR}'22)}, 2022.
\newblock Published online: \url{iclr.cc}.

\bibitem[Haarnoja et~al.(2018)Haarnoja, Zhou, Abbeel, and Levine]{haarnoja-icml18a}
T.~Haarnoja, A.~Zhou, P.~Abbeel, and S.~Levine.
\newblock Soft actor-critic: Off-policy maximum entropy deep reinforcement learning with a stochastic actor.
\newblock In J.~Dy and A.~Krause (eds.), \emph{Proceedings of the 35th International Conference on Machine Learning ({ICML}'18)}, volume~80. Proceedings of Machine Learning Research, 2018.

\bibitem[Hallak et~al.(2015)Hallak, Castro, and Mannor]{hallak-arxiv15a}
A.~Hallak, D.~Di Castro, and S.~Mannor.
\newblock Contextual markov decision processes.
\newblock \emph{arXiv:1502.02259 {[stat.ML]}}, 2015.

\bibitem[Henderson et~al.(2018)Henderson, Islam, Bachman, Pineau, Precup, and Meger]{henderson-aaai18a}
P.~Henderson, R.~Islam, P.~Bachman, J.~Pineau, D.~Precup, and D.~Meger.
\newblock Deep reinforcement learning that matters.
\newblock In S.~McIlraith and K.~Weinberger (eds.), \emph{Proceedings of the Thirty-Second Conference on Artificial Intelligence ({AAAI}'18)}. {AAAI} Press, 2018.

\bibitem[Kaddour et~al.(2020)Kaddour, S{\ae}mundsson, and Deisenroth]{kaddour-neurips20a}
J.~Kaddour, S.~S{\ae}mundsson, and M.~P. Deisenroth.
\newblock Probabilistic active meta-learning.
\newblock In H.~Larochelle, M.~Ranzato, R.~Hadsell, M.-F. Balcan, and H.~Lin (eds.), \emph{Proceedings of the 34th International Conference on Advances in Neural Information Processing Systems ({N}eur{IPS}'20)}. Curran Associates, 2020.

\bibitem[Kirk et~al.(2023)Kirk, Zhang, Grefenstette, and Rockt{\"a}schel]{kirk-jair23a}
R.~Kirk, A.~Zhang, E.~Grefenstette, and T.~Rockt{\"a}schel.
\newblock A survey of zero-shot generalisation in deep reinforcement learning.
\newblock \emph{Journal of Artificial Intelligence Research (JAIR)}, 76:\penalty0 201--264, 2023.

\bibitem[Koppejan \& Whiteson(2009)Koppejan and Whiteson]{koppejan-gecco09a}
R.~Koppejan and S.~Whiteson.
\newblock Neuroevolutionary reinforcement learning for generalized helicopter control.
\newblock In G.~Raidl et~al (ed.), \emph{Proceedings of the 11th Genetic and Evolutionary Computation Conference ({GECCO}'09)}. Morgan Kaufmann Publishers, 2009.

\bibitem[Lee et~al.(2020)Lee, Seo, Lee, Lee, and Shin]{lee-icml20b}
K.~Lee, Y.~Seo, S.~Lee, H.~Lee, and J.~Shin.
\newblock Context-aware dynamics model for generalization in model-based reinforcement learning.
\newblock In H.~{Daume III} and A.~Singh (eds.), \emph{Proceedings of the 37th International Conference on Machine Learning ({ICML}'20)}, volume~98, pp.\  5757--5766. Proceedings of Machine Learning Research, 2020.

\bibitem[Melo(2022)]{melo-icml22a}
L.~C. Melo.
\newblock Transformers are meta-reinforcement learners.
\newblock In K.~Chaudhuri, S.~Jegelka, L.~Song, C.~Szepesvári, G.~Niu, and S.~Sabato (eds.), \emph{Proceedings of the 39th International Conference on Machine Learning ({ICML}'22)}, volume 162 of \emph{Proceedings of Machine Learning Research}, pp.\  15340--15359. PMLR, 2022.

\bibitem[Modi et~al.(2018)Modi, Jiang, Singh, and Tewari]{modi-alt18a}
A.~Modi, N.~Jiang, S.~Singh, and A.~Tewari.
\newblock Markov decision processes with continuous side information.
\newblock In \emph{Algorithmic Learning Theory ({ALT}'18)}, volume~83, pp.\  597--618, 2018.

\bibitem[Nagabandi et~al.(2019)Nagabandi, Clavera, Liu, Fearing, Abbeel, Levine, and Finn]{nagabandi-iclr19a}
A.~Nagabandi, I.~Clavera, S.~Liu, R.~S. Fearing, P.~Abbeel, S.~Levine, and C.~Finn.
\newblock Learning to adapt in dynamic, real-world environments through meta-reinforcement learning.
\newblock In \emph{Proceedings of the International Conference on Learning Representations ({ICLR}'19)}, 2019.
\newblock Published online: \url{iclr.cc}.

\bibitem[Parker-Holder et~al.(2022)Parker-Holder, Rajan, Song, Biedenkapp, Miao, Eimer, Zhang, Nguyen, Calandra, Faust, Hutter, and Lindauer]{parkerholder-jair22a}
J.~Parker-Holder, R.~Rajan, X.~Song, A.~Biedenkapp, Y.~Miao, T.~Eimer, B.~Zhang, V.~Nguyen, R.~Calandra, A.~Faust, F.~Hutter, and M.~Lindauer.
\newblock Automated reinforcement learning ({A}uto{RL}): A survey and open problems.
\newblock \emph{Journal of Artificial Intelligence Research (JAIR)}, 74:\penalty0 517--568, 2022.

\bibitem[Peng et~al.(2018)Peng, Andrychowicz, Zaremba, and Abbeel]{peng-icra18a}
X.~B. Peng, M.~Andrychowicz, W.~Zaremba, and P.~Abbeel.
\newblock Sim-to-real transfer of robotic control with dynamics randomization.
\newblock In \emph{International Conference on Robotics and Automation, ({ICRA}'18)}, pp.\  1--8. {IEEE}, 2018.

\bibitem[Rakelly et~al.(2019)Rakelly, Zhou, Finn, Levine, and Quillen]{rakelly_efficient_2019}
K.~Rakelly, A.~Zhou, C.~Finn, S.~Levine, and D.~Quillen.
\newblock Efficient off-policy meta-reinforcement learning via probabilistic context variables.
\newblock In K.~Chaudhuri and R.~Salakhutdinov (eds.), \emph{Proceedings of the 36th International Conference on Machine Learning ({ICML}'19)}, volume~97, pp.\  5331--5340. Proceedings of Machine Learning Research, 2019.

\bibitem[Schulman et~al.(2017)Schulman, Wolski, Dhariwal, Radford, and Klimov]{schulman-arxiv17a}
J.~Schulman, F.~Wolski, P.~Dhariwal, A.~Radford, and O.~Klimov.
\newblock Proximal policy optimization algorithms.
\newblock \emph{arXiv:1707.06347 [cs.LG]}, 2017.

\bibitem[Seo et~al.(2020)Seo, Lee, Gilaberte, Kurutach, Shin, and Abbeel]{seo-neurips20a}
Y.~Seo, K.~Lee, I.~C. Gilaberte, T.~Kurutach, J.~Shin, and P.~Abbeel.
\newblock Trajectory-wise multiple choice learning for dynamics generalization in reinforcement learning.
\newblock In H.~Larochelle, M.~Ranzato, R.~Hadsell, M.-F. Balcan, and H.~Lin (eds.), \emph{Proceedings of the 34th International Conference on Advances in Neural Information Processing Systems ({N}eur{IPS}'20)}. Curran Associates, 2020.

\bibitem[Sodhani et~al.(2022)Sodhani, Meier, Pineau, and Zhang]{sodhani-ldcc22a}
S.~Sodhani, F.~Meier, J.~Pineau, and A.~Zhang.
\newblock Block contextual mdps for continual learning.
\newblock In R.~Firoozi, N.~Mehr, E.~Yel, R.~Antonova, J.~Bohg, M.~Schwager, and M.~J. Kochenderfer (eds.), \emph{Learning for Dynamics and Control Conference, ({L4DC}'22)}, volume 168 of \emph{Proceedings of Machine Learning Research}, pp.\  608--623. {PMLR}, 2022.

\bibitem[Sutton \& Barto(2018)Sutton and Barto]{sutton-book18a}
R.~S. Sutton and A.~G. Barto.
\newblock \emph{Reinforcement learning: An introduction}.
\newblock Adaptive computation and machine learning. {MIT} Press, 2 edition, 2018.

\bibitem[Tobin et~al.(2017)Tobin, Fong, Ray, Schneider, Zaremba, and Abbeel]{tobin-iros17a}
J.~Tobin, R.~Fong, A.~Ray, J.~Schneider, W.~Zaremba, and P.~Abbeel.
\newblock Domain randomization for transferring deep neural networks from simulation to the real world.
\newblock In \emph{International Conference on Intelligent Robots and Systems ({IROS}'17)}, pp.\  23--30, 2017.

\bibitem[Wang et~al.(2017)Wang, Kurth{-}Nelson, Soyer, Leibo, Tirumala, Munos, Blundell, Kumaran, and Botvinick]{wang-cogsci17a}
J.~Wang, Z.~Kurth{-}Nelson, H.~Soyer, J.~Leibo, D.~Tirumala, R.~Munos, C.~Blundell, D.~Kumaran, and M.~Botvinick.
\newblock Learning to reinforcement learn.
\newblock In G.~Gunzelmann, A.~Howes, T.~Tenbrink, and E.~Davelaar (eds.), \emph{Proceedings of the 39th Annual Meeting of the Cognitive Science Society}. cognitivesciencesociety.org, 2017.

\bibitem[Wen et~al.(2023)Wen, Zhang, Tseng, and Peng]{wen-arxiv23a}
L.~Wen, S.~Zhang, H.~E. Tseng, and H.~Peng.
\newblock Dream to adapt: Meta reinforcement learning by latent context imagination and {MDP} imagination.
\newblock \emph{arXiv:2311.06673 [cs.LG]}, 2023.

\bibitem[Zhou et~al.(2019)Zhou, Pinto, and Gupta]{zhou_environment_2022}
W.~Zhou, L.~Pinto, and A.~Gupta.
\newblock Environment probing interaction policies.
\newblock In \emph{Proceedings of the International Conference on Learning Representations ({ICLR}'19)}, 2019.
\newblock Published online: \url{iclr.cc}.

\end{thebibliography}
\bibliographystyle{rlc}

%%%%%%%%%%%%%%%%%%%%%%%%%%%%%%%%%%%%%%%%%%%%%%%%%%%%%%%%%%%%%%%%
%% Appendices
%%%%%%%%%%%%%%%%%%%%%%%%%%%%%%%%%%%%%%%%%%%%%%%%%%%%%%%%%%%%%%%%
\appendix

%\section{The first appendix}
%\label{sec:appendix1}
%This is an example of an appendix.

%\section{The second appendix}
%\label{sec:appendix2}
%This is an example of a second appendix. If there is only a single section in the appendix, you may simply call it ``Appendix'' as follows:

\section*{Appendix}
% No label, since this can't be referenced meaningfully with \ref{}.
%This format should only be used if there is a single appendix (unlike in this document).

\begin{table}[h]
    \centering
    \begin{tabular}{l|l}
    \toprule
    \textbf{Hyperparameter} & \textbf{Value} \\
    \midrule
    Total number of training steps $T_{train}$ & $3 \times 10^{4}$ \\ 
    Replay Buffer Size & $10^6$ \\
    Discount Factor $\gamma$ & 0.99 \\
    Polyak Averaging Factor $\tau$ & 0.005 \\
    Batch Size b & 256 \\
    Time steps before training starts  & $5 \times 10^{3}$ \\
    Learning Rate (Actor) & $3 \times 10^{-4}$ \\
    Learning Rate (Critic) & $10^{-3}$ \\   
    Optimizer (Actor, Critic) & Adam \\
    Policy training interval & 2 \\
    Target Update Interval & 1 \\
    Target Entropy $\alpha_{\text{target}}$ & -1 \\
    \bottomrule
    \end{tabular}
    \caption{Hyperparameters relative to the task policy algorithm (Soft Actor-Critic)}
    \label{tab:sac_hyperparameters}
\end{table}

\begin{table}[h]
    \centering
    \begin{tabular}{l|l}
    \toprule
    \textbf{Hyperparameter} & \textbf{Value} \\
    \midrule
    Size $h$ of transitions list $L$ & 20 \\
    Dimension of latent context $l_c$ & 2 \\
    Hidden dimensions (Encoder $\psi$, Predictor $D$) & [8,4] \\
    Learning rate (Encoder $\psi$, Predictor $D$) & $10^{-3}$  \\
    Optimizer (Encoder $\psi$, Predictor $D$) & Adam  \\
    \bottomrule
    \end{tabular}
    \caption{Hyperparameters relative to the encoding of the context}
    \label{tab:encoder_hyperparameters}
\end{table}

\begin{figure}[htbp!]
    \centering
    \begin{subfigure}[b]{\textwidth}
    \includegraphics[width=\textwidth]{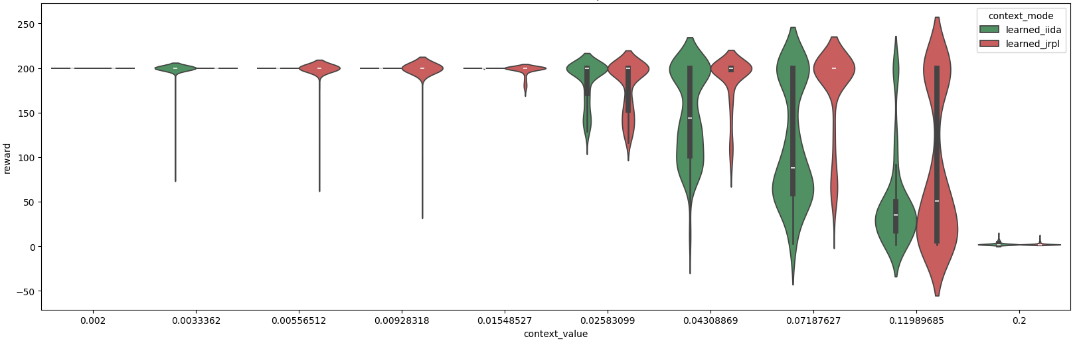}
    \caption{Cartpole}
    \label{fig:exp-zsg-cartpole}
    \end{subfigure}
    \begin{subfigure}[b]{\textwidth}
    \includegraphics[width=\textwidth]{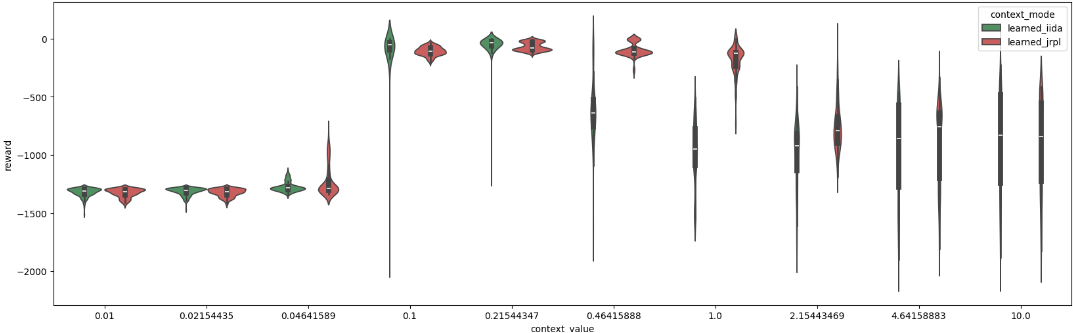}
    \caption{Pendulum}
    \label{fig:exp-zsg-pendulum}
    \end{subfigure}
    \caption{Violin plots of the (non-normalized) scores for every value of $C_{eval}$, in the Cartpole and Pendulum environments. Our joint learning method (jcpl) is capable of zero-shot generalization to out-of-distribution environment dynamics.  }
    \label{fig:exp-comparison}
\end{figure}

\begin{figure}[htbp!]
    \centering
    \begin{subfigure}[b]{.95\textwidth}
    \includegraphics[width=\textwidth]{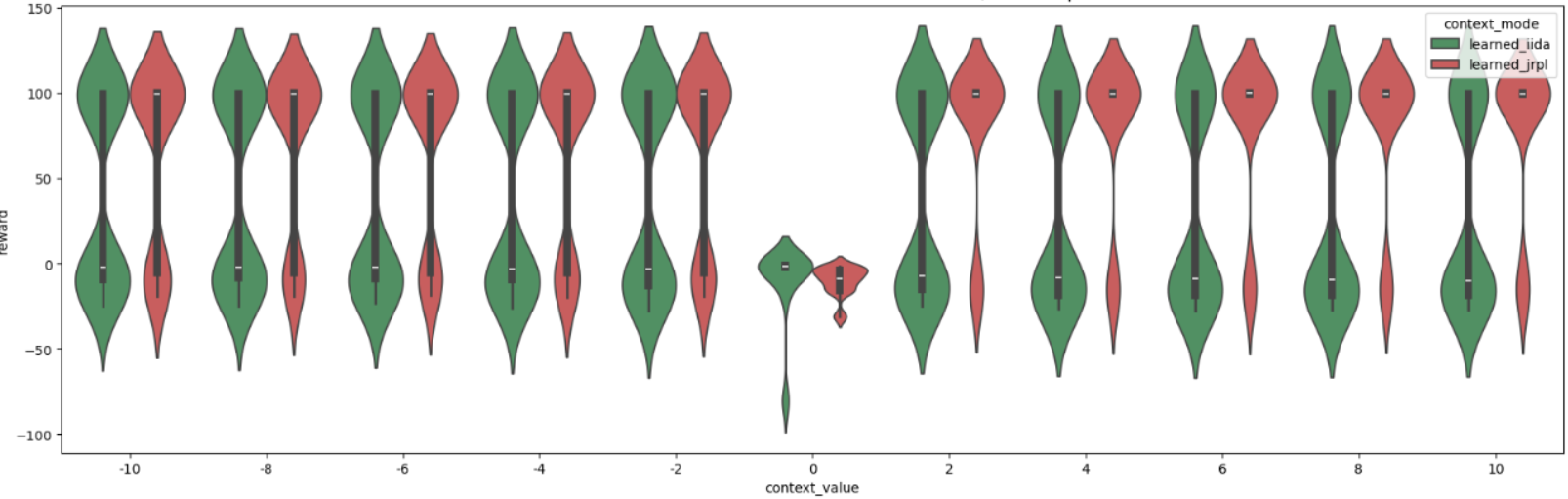}
    \caption{MountainCar}
    \label{fig:exp-zsg-mc}
    \end{subfigure}
    \begin{subfigure}[b]{.95\textwidth}
    \includegraphics[width=\textwidth]{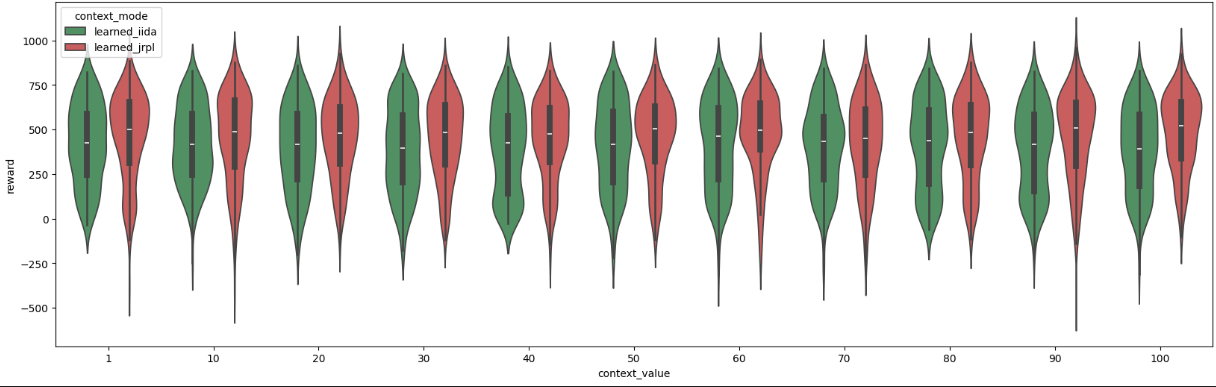}
    \caption{Ant}
    \label{fig:exp-zsg-ant}
    \end{subfigure}
    \caption{Violin plots of the (non-normalized) scores for every value of $C_{eval}$, in the MountainCar and Ant environments. }
    \label{fig:generalization_violin_plots}
\end{figure}

\begin{table}[htbp]
\centering
\begin{tabular}{c|c|c|l}
\toprule
\textbf{Environment ($T_{episode}$)} & \textbf{Context} & \textbf{$C_{train}$} & \textbf{$C_{eval}$}\\
\midrule
\multirow{4}{*}{Cartpole (200)} & \multirow{4}{*}{tau} & [0.007, 0.012, & [0.002, 0.003, \\
& & 0.021, 0.034, & 0.006, 0.009, \\
& & 0.057] & 0.015, 0.026, \\
& & & 0.043, 0.072, \\
& & & 0.120, 0.200] \\
\midrule
\multirow{4}{*}{Pendulum (200)} & \multirow{4}{*}{length} & [0.07, 0.16, & [0.01, 0.02, \\
& & 0.34, 0.73, & 0.05, 0.1, \\
& & 1.58] & 0.22, 0.46, \\
& & & 1.0, 2.15, \\
& & & 4.64, 10.00] \\
\midrule
\multirow{4}{*}{MountainCar (999)} & \multirow{4}{*}{power} & [-5, -3, -1, & [-10, -8, -6, \\
& & 1, 3, 5] & -4, -2, 0, \\
& & & 2, 4, 6, \\
& & & 8, 10] \\
\midrule
\multirow{4}{*}{Ant (1000)} & \multirow{4}{*}{mass\_torso} & [25, 35, 45, & [1, 10, 20, \\
& & 55, 65, 75] & 30, 40, 50, \\
& & & 60, 70, 80, \\
& & & 90, 100] \\
\bottomrule
\end{tabular}
\caption{Environment settings}
\label{tab:environment_context}
\end{table}

\begin{figure}[htbp!]
    \centering

    \begin{subfigure}[b]{0.40\textwidth}
    \includegraphics[width=\textwidth]{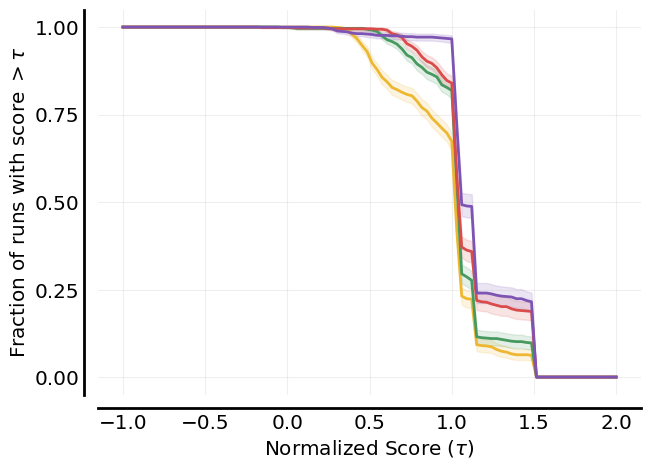}
    \caption{Cartpole - interpolation}
    \label{fig:PP_Cartpole_interpolation}
    \end{subfigure}
    \begin{subfigure}[b]{0.40\textwidth}
    \includegraphics[width=\textwidth]{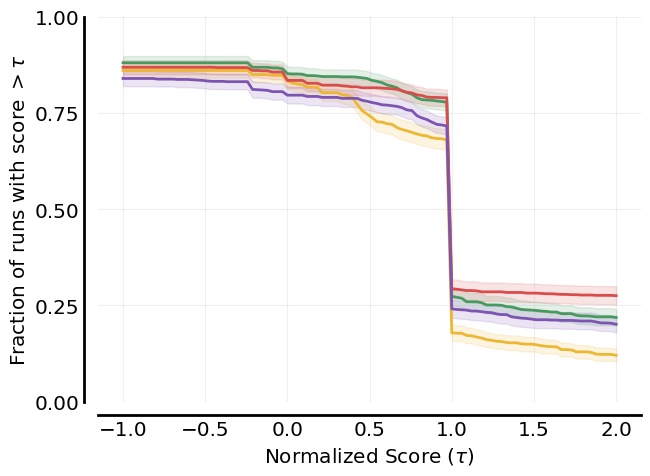}
    \caption{Cartpole - extrapolation}
    \label{fig:cartpole_pp_extrapolation}
    \end{subfigure}
    \\
    \begin{subfigure}[b]{0.40\textwidth}
    \includegraphics[width=\textwidth]{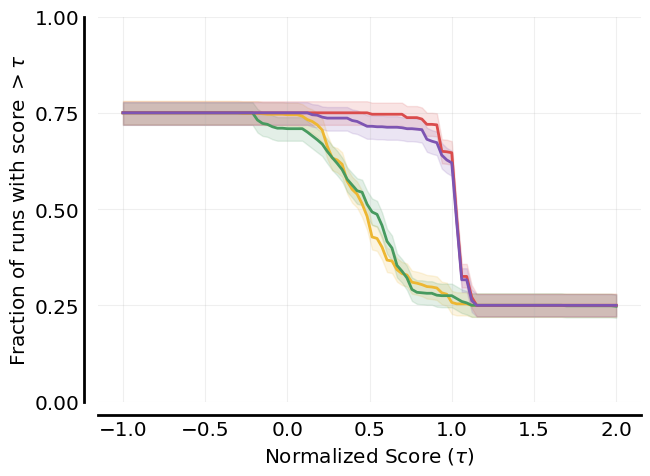}
    \caption{Pendulum - interpolation}
    \label{fig:pendulum_pp_interpolation}
    \end{subfigure}
    \begin{subfigure}[b]{0.40\textwidth}
    \includegraphics[width=\textwidth]{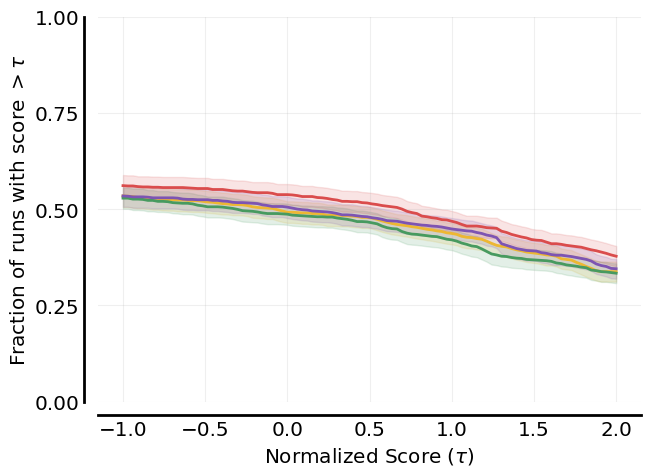}
    \caption{Pendulum - extrapolation}
    \label{fig:exp-comparison-B}
    \end{subfigure}
    \\
    \begin{subfigure}[b]{0.40\textwidth}
    \includegraphics[width=\textwidth]{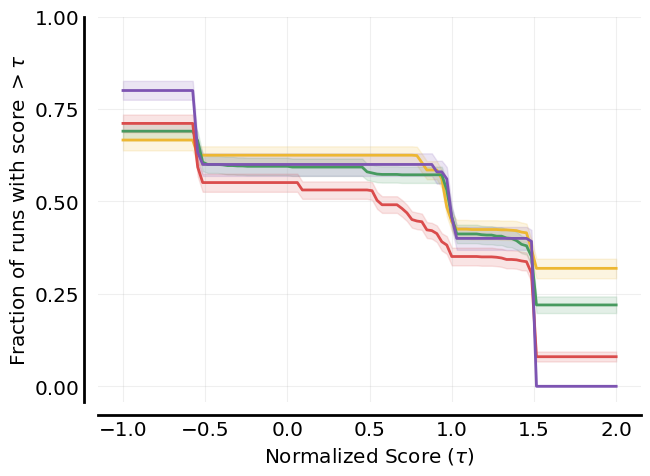}
    \caption{MountainCar - interpolation}
    \label{fig:mountaincar_pp_interpolation}
    \end{subfigure}
    \begin{subfigure}[b]{0.40\textwidth}
    \includegraphics[width=\textwidth]{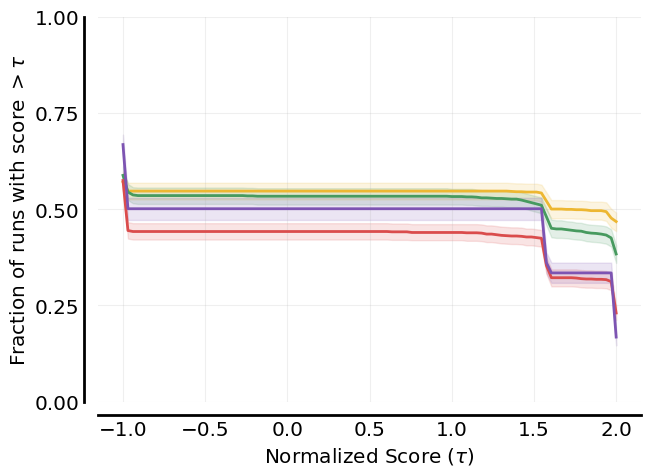}
    \caption{MountainCar - extrapolation}
    \label{fig:mountaincar_pp_extrapolation}
    \end{subfigure}
    \\
    \begin{subfigure}[b]{0.40\textwidth}
    \includegraphics[width=\textwidth]{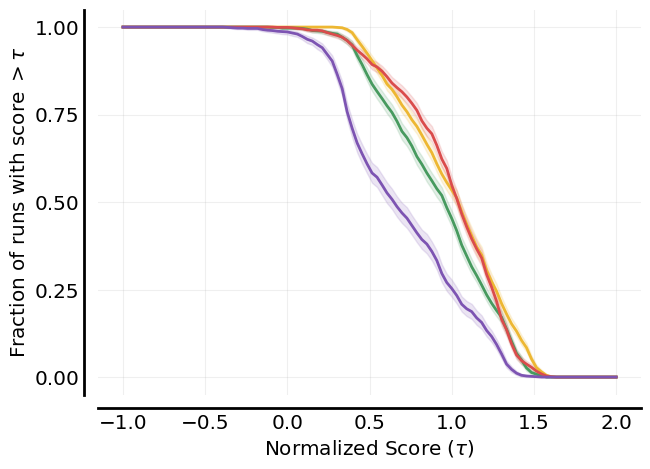}
    \caption{Ant - interpolation}
    \label{fig:PP_Ant_Intra}
    \end{subfigure}
    \begin{subfigure}[b]{0.40\textwidth}
    \includegraphics[width=\textwidth]{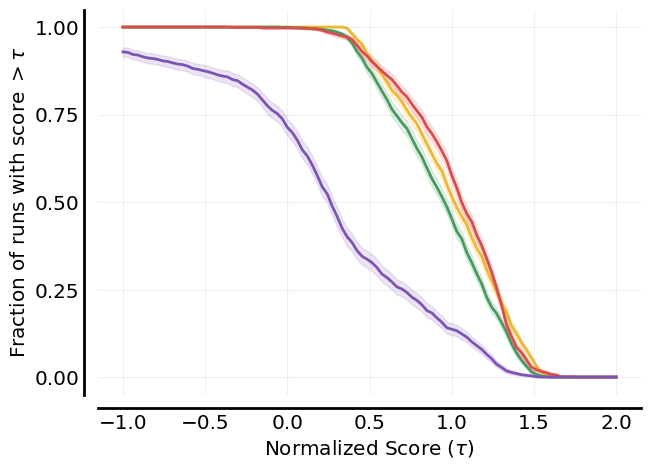}
    \caption{Ant - extrapolation}
    \label{fig:PP_Ant_extra}
    \end{subfigure}

    \caption{Performance profiles of aggregated normalized scores, in the interpolation and extrapolation settings}
    \label{fig:generalization-metrics-intra-extra}
\end{figure}

\begin{figure}[htbp!]
    \centering
    \begin{subfigure}[b]{0.40\textwidth}
    \includegraphics[width=\textwidth]{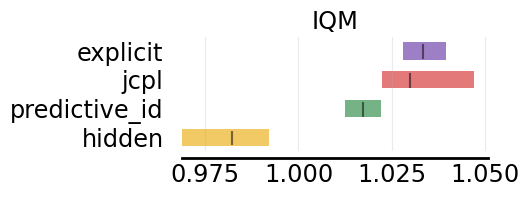}
    \caption{Cartpole - IQM of aggregated normalized scores}
    \label{fig:exp-comparison-C}
    \end{subfigure}
    \begin{subfigure}[b]{0.40\textwidth}
    \includegraphics[width=\textwidth]{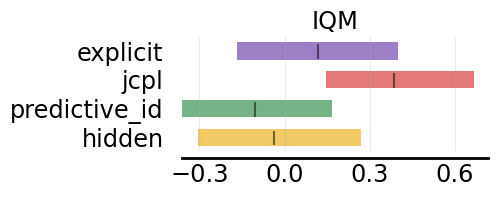}
    \caption{Pendulum - IQM of aggregated normalized scores}
    \label{fig:IQM_Pendulum}
    \end{subfigure}
    \\
    \begin{subfigure}[b]{0.40\textwidth}
    \includegraphics[width=\textwidth]{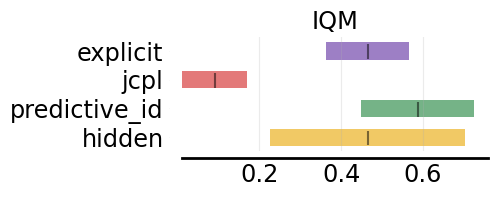}
    \caption{MountainCar - IQM of aggregated normalized scores}
    \label{fig:IQM_MountainCar}
    \end{subfigure}
    \begin{subfigure}[b]{0.40\textwidth}
    \includegraphics[width=\textwidth]{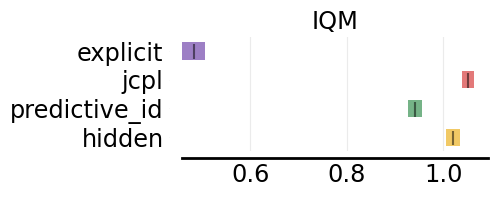}
    \caption{Ant - IQM of aggregated normalized scores}
    \label{fig:IQM_Ant}
    \end{subfigure}
    \\
    \begin{subfigure}[b]{0.40\textwidth}
    \includegraphics[width=\textwidth]{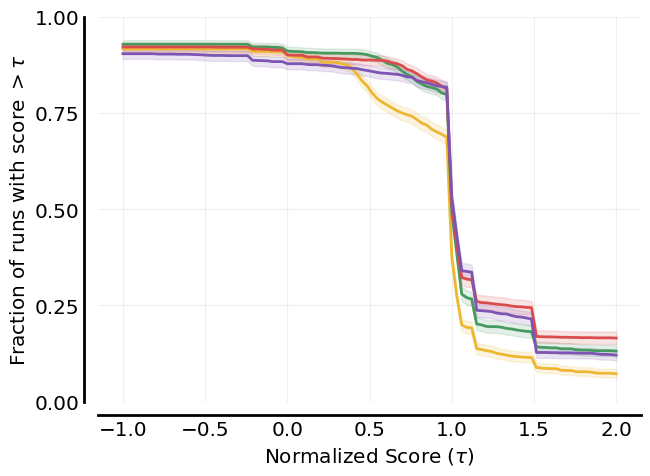}
    \caption{Cartpole - performance profiles of aggregated normalized scores}
    \label{fig:PP_Cartpole}
    \end{subfigure}
    \begin{subfigure}[b]{0.40\textwidth}performance
    \includegraphics[width=\textwidth]{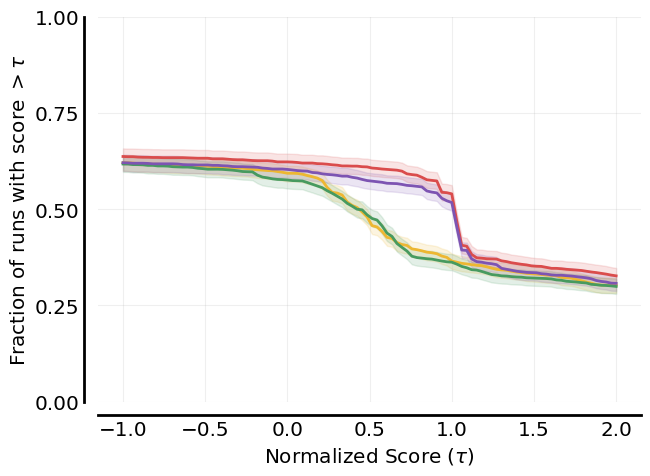}
    \caption{Pendulum - performance profiles of aggregated normalized scores}
    \label{fig:PP_Pendulum}
    \end{subfigure}
    \\
    \begin{subfigure}[b]{0.40\textwidth}
    \includegraphics[width=\textwidth]{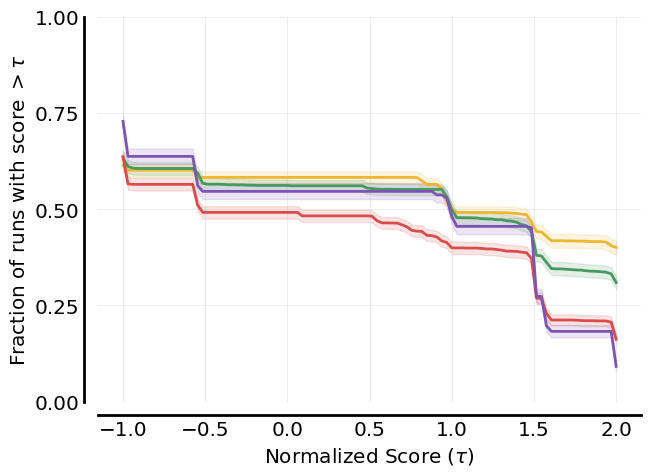}
    \caption{MountainCar - performance profiles of aggregated normalized scores}
    \label{fig:PP_Mountaincar}
    \end{subfigure}
    \begin{subfigure}[b]{0.40\textwidth}
    \includegraphics[width=\textwidth]{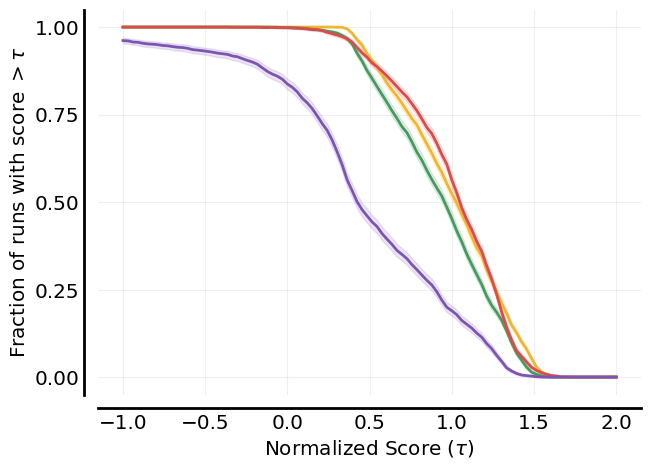}
    \caption{Ant - performance profiles of aggregated normalized scores}
    \label{fig:PP_Ant}
    \end{subfigure}
    \caption{Interquartile Mean (IQM) and performance profiles of aggregated normalized scores, on the entire $C_{eval}$ set}
    \label{fig:generalization-metrics-all}
\end{figure}

\begin{figure}[htbp!]
    \centering
    
    \begin{subfigure}[b]{.24\textwidth}
    \includegraphics[width=\textwidth]{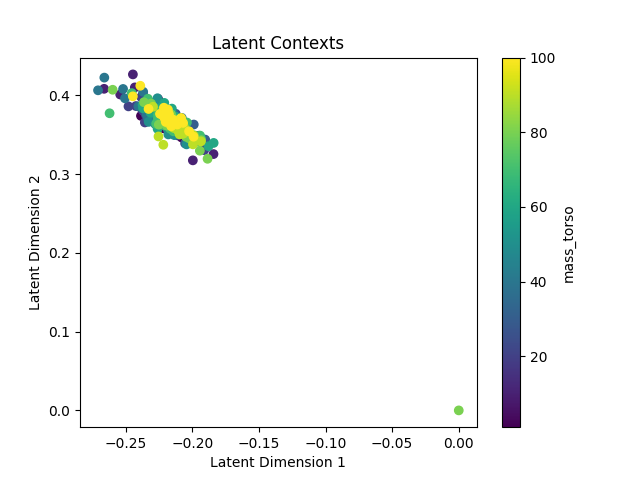}
    \caption{Ant - predictive identification}
    \end{subfigure}
    \begin{subfigure}[b]{.24\textwidth}
    \includegraphics[width=\textwidth]{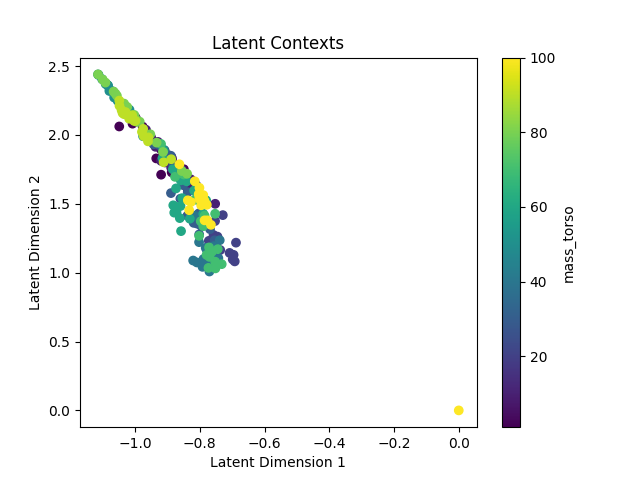}
    \caption{Ant - jcpl}
    \end{subfigure}
    \begin{subfigure}[b]{.24\textwidth}
    \includegraphics[width=\textwidth]{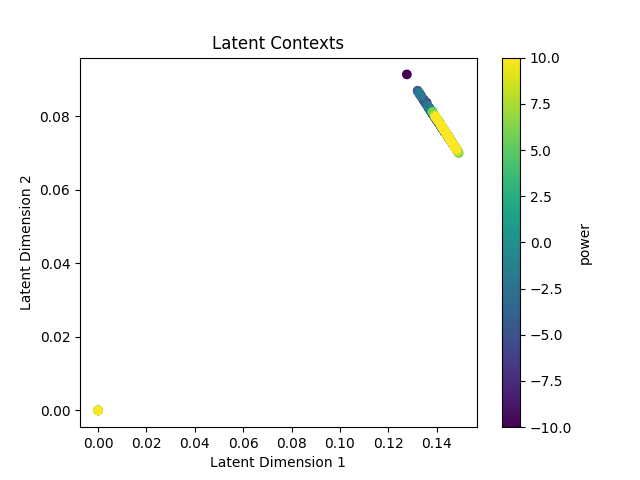}
    \caption{MountainCar - predictive identification}
    \end{subfigure}
    \begin{subfigure}[b]{.24\textwidth}
    \includegraphics[width=\textwidth]{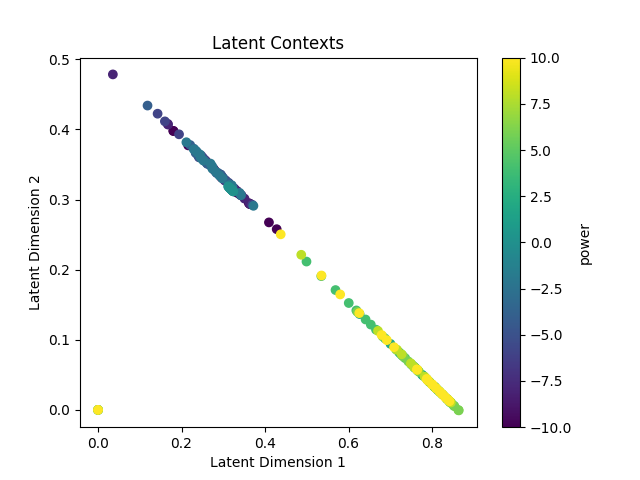}
    \caption{MountainCar - jcpl}
    \end{subfigure}
    
    \vspace{0.5cm}
    
    \begin{subfigure}[b]{.24\textwidth}
    \includegraphics[width=\textwidth]{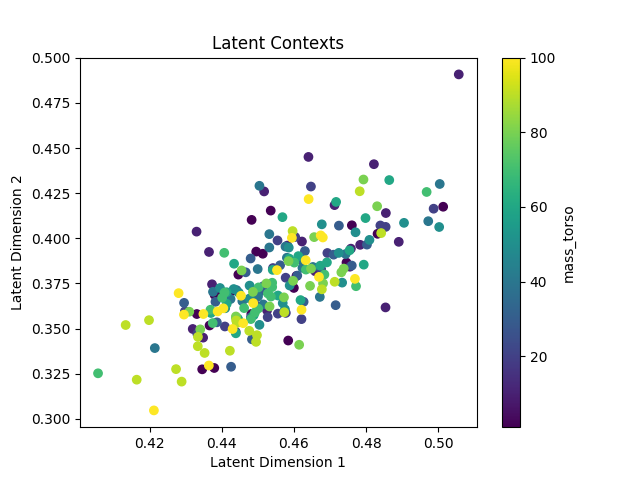}
    \caption{Ant - predictive identification}
    \end{subfigure}
    \begin{subfigure}[b]{.24\textwidth}
    \includegraphics[width=\textwidth]{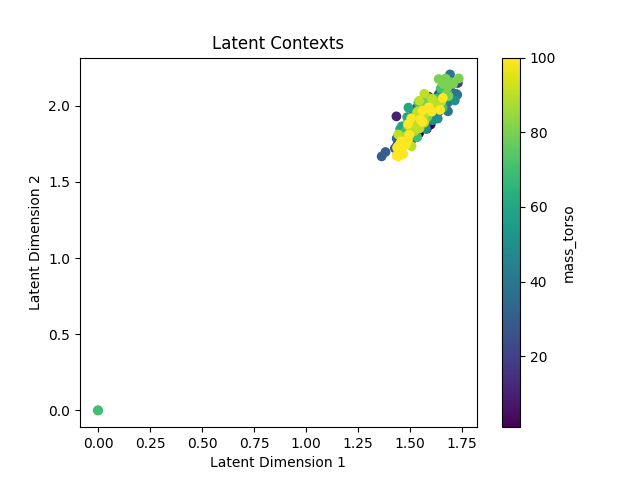}
    \caption{Ant - jcpl}
    \end{subfigure}
    \begin{subfigure}[b]{.24\textwidth}
    \includegraphics[width=\textwidth]{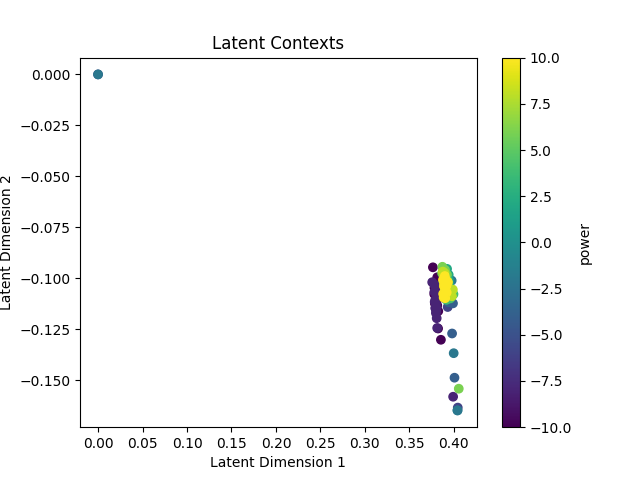}
    \caption{MountainCar - predictive identification}
    \end{subfigure}
    \begin{subfigure}[b]{.24\textwidth}
    \includegraphics[width=\textwidth]{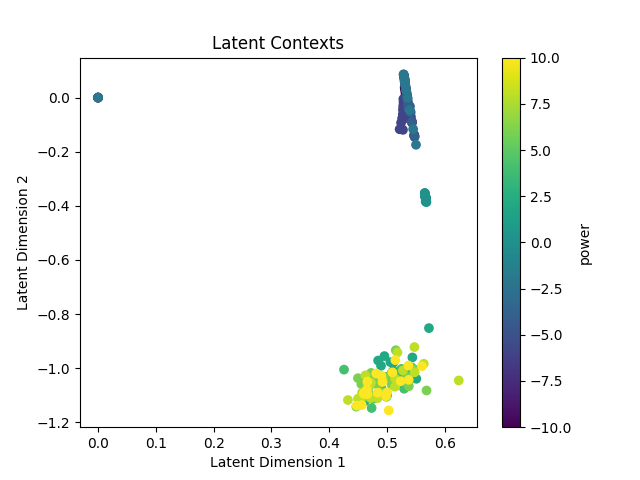}
    \caption{MountainCar - jcpl}
    \end{subfigure}
    
    \vspace{0.5cm} % adjust vertical space between rows
    
    \vspace{0.5cm} % adjust vertical space between rows
    
    \begin{subfigure}[b]{.24\textwidth}
    \includegraphics[width=\textwidth]{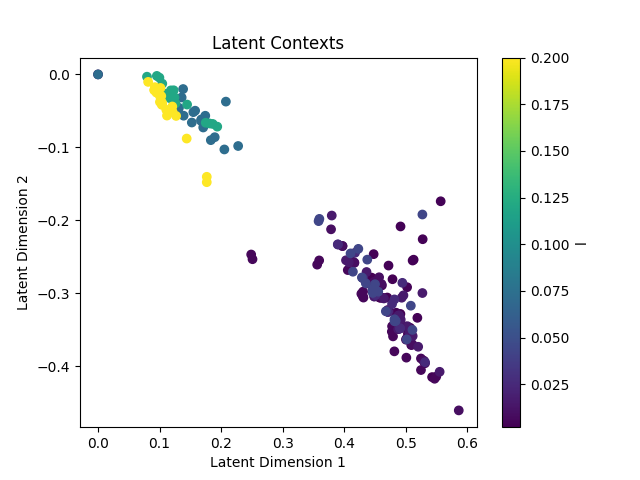}
    \caption{Pendulum - predictive identification}
    \end{subfigure}
    \begin{subfigure}[b]{.24\textwidth}
    \includegraphics[width=\textwidth]{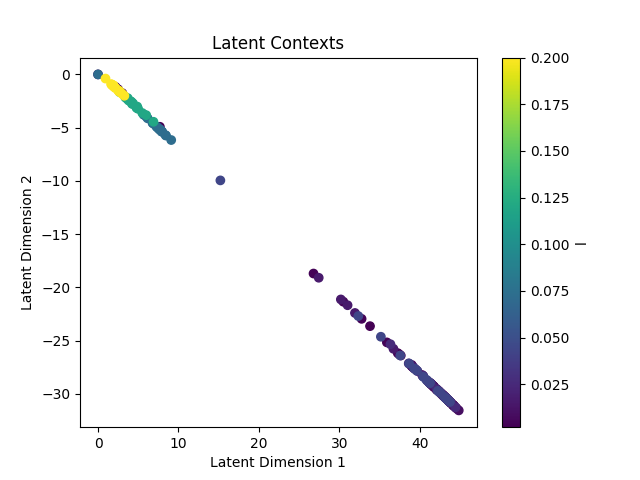}
    \caption{Pendulum - jcpl}
    \end{subfigure}
    \begin{subfigure}[b]{.24\textwidth}
    \includegraphics[width=\textwidth]{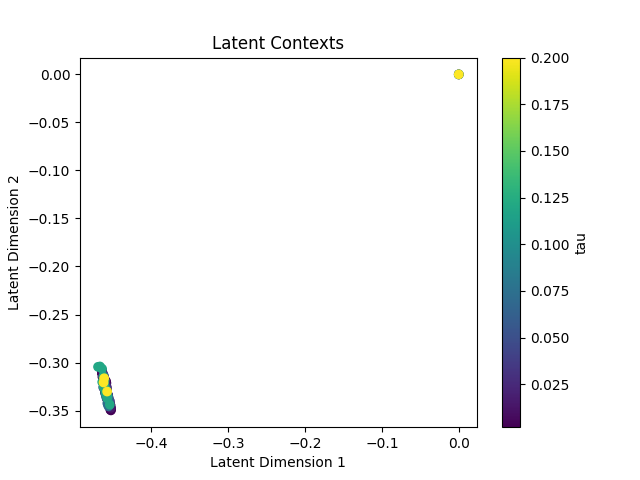}
    \caption{Cartpole - predictive identification}
    \end{subfigure}
    \begin{subfigure}[b]{.24\textwidth}
    \includegraphics[width=\textwidth]{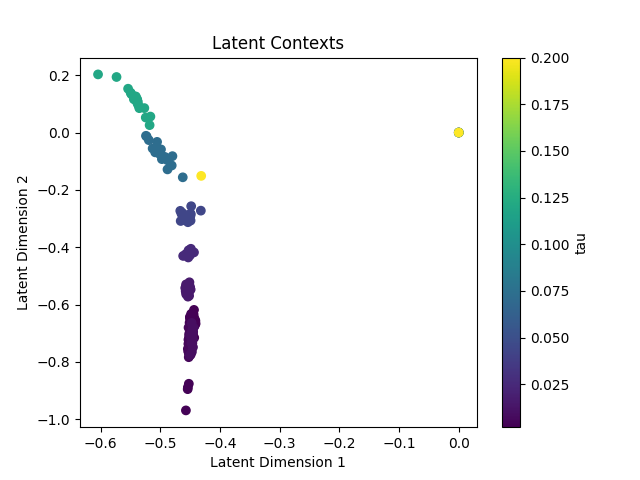}
    \caption{Cartpole - jcpl}
    \end{subfigure}
    
    \vspace{0.5cm} % adjust vertical space between rows    \vspace{0.5cm} % adjust vertical space between rows
    
    \begin{subfigure}[b]{.24\textwidth}
    \includegraphics[width=\textwidth]{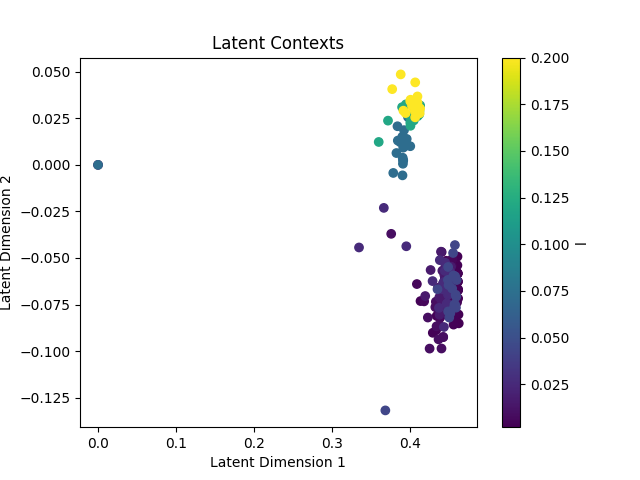}
    \caption{Pendulum - predictive identification}
    \end{subfigure}
    \begin{subfigure}[b]{.24\textwidth}
    \includegraphics[width=\textwidth]{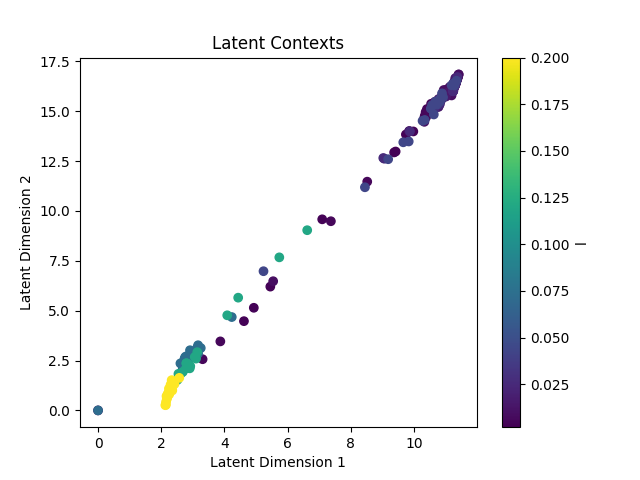}
    \caption{Pendulum - jcpl}
    \end{subfigure}
    \begin{subfigure}[b]{.24\textwidth}
    \includegraphics[width=\textwidth]{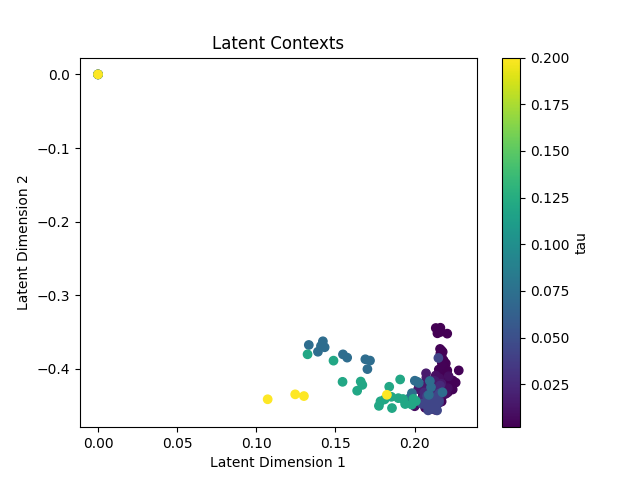}
    \caption{Cartpole - predictive identification}
    \end{subfigure}
    \begin{subfigure}[b]{.24\textwidth}
    \includegraphics[width=\textwidth]{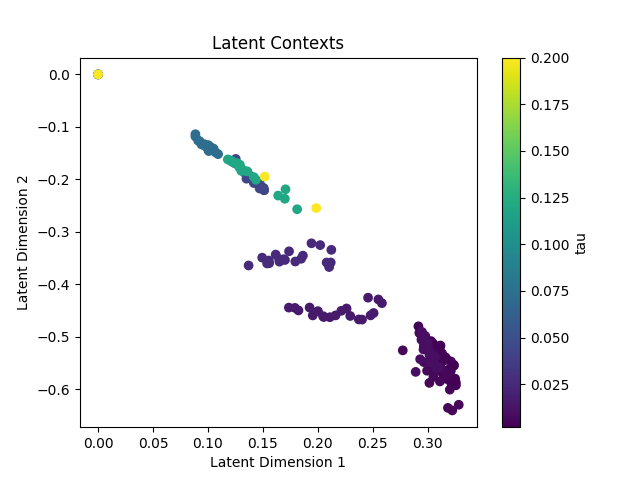}
    \caption{Cartpole - jcpl}
    \end{subfigure}

    \vspace{0.5cm} % adjust vertical space between rows
    
    \begin{subfigure}[b]{.24\textwidth}
    \includegraphics[width=\textwidth]{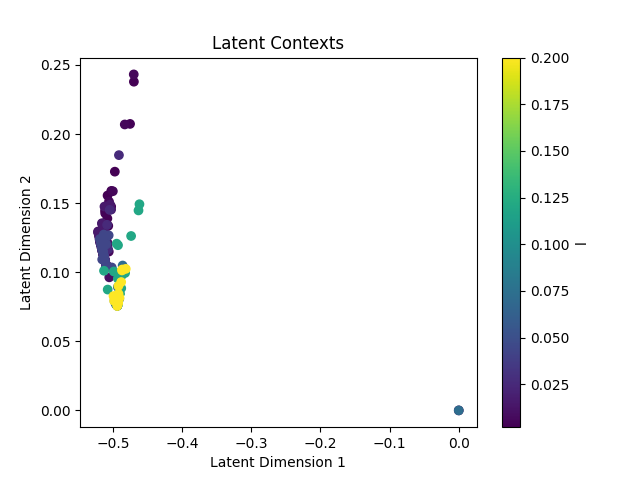}
    \caption{Pendulum - predictive identification}
    \end{subfigure}
    \begin{subfigure}[b]{.24\textwidth}
    \includegraphics[width=\textwidth]{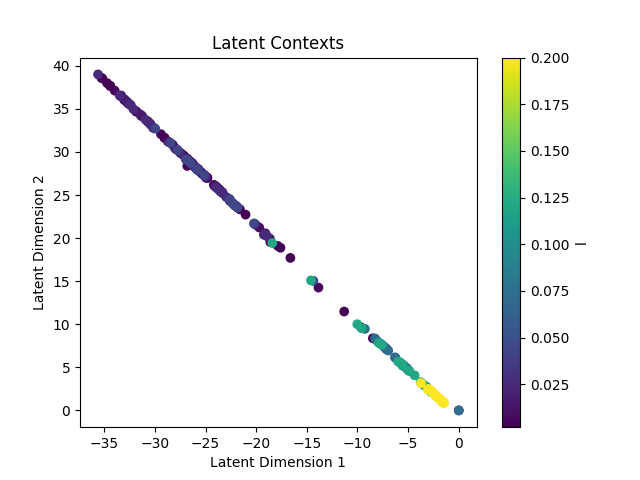}
    \caption{Pendulum - jcpl}
    \end{subfigure}
    \begin{subfigure}[b]{.24\textwidth}
    \includegraphics[width=\textwidth]{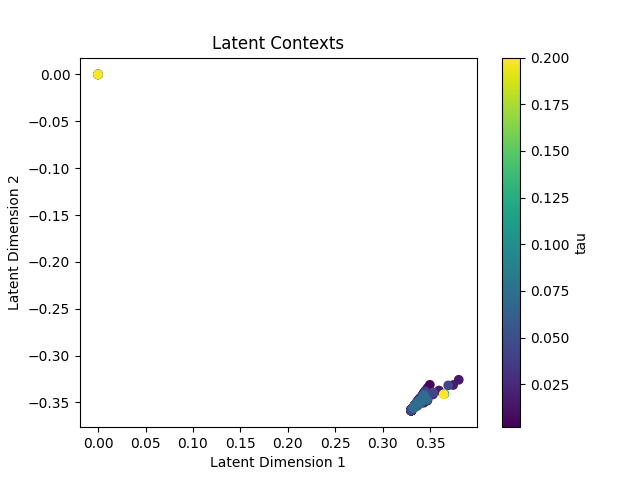}
    \caption{Cartpole - predictive identification}
    \end{subfigure}
    \begin{subfigure}[b]{.24\textwidth}
    \includegraphics[width=\textwidth]{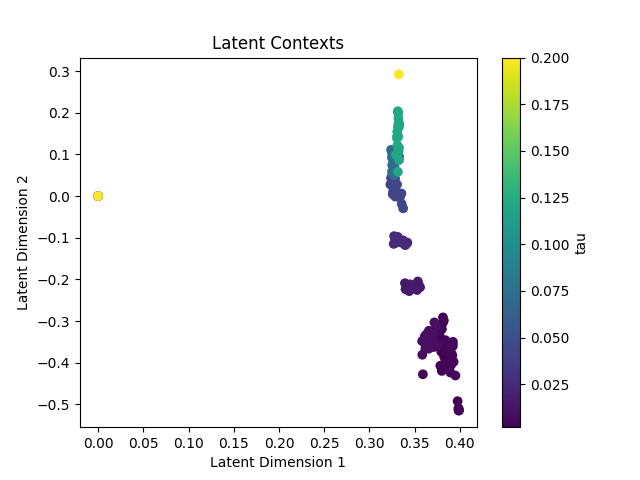}
    \caption{Cartpole - jcpl}
    \end{subfigure}

    \caption{Learned embeddings of both context encoding methods, across multiple seeds}
    \label{fig:latent-comparison}
\end{figure}

\end{document}